\documentclass[fleqn]{article}
\usepackage{graphicx,color,wallpaper,amssymb,enumitem}
\usepackage{algorithm} 
\usepackage{algpseudocode} 
\usepackage{authblk,wrapfig}

\usepackage{hyperref}
\hypersetup{
    colorlinks=true,
    linkcolor=blue,
    filecolor=magenta,      
    urlcolor=cyan,
}

\usepackage{amssymb,array,delarray,verbatim,alltt,epsfig,float,amsthm}
\usepackage{graphicx,algorithmicx,algorithm,algpseudocode,xspace,xcolor}
\usepackage{url}
\usepackage{epsf,epsfig,subfigure,dsfont}
\usepackage{amssymb,amsmath}
\usepackage{pgf,amsbsy,amsfonts,amsmath,subfigure,multicol}
\usepackage{graphicx,xspace,color,cancel}
\usepackage{tikz}
\usetikzlibrary{arrows,shapes,backgrounds,through,shadows}
\usetikzlibrary{decorations.pathmorphing}
\usetikzlibrary{calc}
\usetikzlibrary{matrix}

\graphicspath{{./images/}{./matlab/}{./figures/}{../matlab/}{../matlab/figures/}{../cup_first_edition/figures/}{./../codeOO/misc/}{../cup_first_edition_corrections/figures/}{./../codeOO/DemosExercises/}{../../projects/deepbelief/}
{../../talks/CambSciFest/figures/}{../../talks/CambSciFest/images/}}

\tikzstyle{bigdot}=[circle,draw=blue!80,fill=blue!25,anchor=center,align=center,minimum size=6mm]
\tikzstyle{bigemptydot}=[circle,draw=blue!80,anchor=center,align=center,minimum size=6mm]

\tikzstyle{biggreendot}=[circle,draw=green!80,fill=green!25,anchor=center,align=center,minimum size=6mm]
\tikzstyle{bigreddot}=[circle,draw=red!80,fill=red!25,anchor=center,align=center,minimum size=6mm]

\tikzstyle{dot}=[circle,draw=orange!80,fill=orange!25,anchor=center,align=center,minimum size=4mm]
\tikzstyle{emptydot}=[circle,draw=orange!80,anchor=center,align=center,minimum size=4mm]

\tikzstyle{reddot}=[circle,draw=red!80,fill=red!25,anchor=center,align=center,minimum size=4mm]

\tikzstyle{greendot}=[circle,draw=green!80,fill=green!25,anchor=center,align=center,minimum size=4mm]

\tikzstyle{bluedot}=[circle,draw=blue!80,fill=blue!25,anchor=center,align=center,minimum size=4mm]

\tikzstyle{neur}=[rectangle,draw=green!50,fill=green!50,minimum size=6mm,line width=2pt,>=stealth]  

\tikzstyle{fact}=[fill,minimum size=1.5mm,line width=2pt,>=stealth]

\tikzstyle{cont2}=[circle,draw=black!50,top color=white, 
bottom color=lilla!50!black!20,thick,minimum size=6mm,line width=1pt,>=stealth]  

\tikzstyle{contredb}=[circle,draw=red,top color=red, 
bottom color=red!50!black!20,thick,minimum size=8.1mm,line width=1pt,>=stealth]  
\tikzstyle{contyellowb}=[circle,draw=yellow,top color=yellow, 
bottom color=yellow!50!black!20,thick,minimum size=8.1mm,line width=1pt,>=stealth]  
\tikzstyle{contblueb}=[circle,draw=blue,top color=blue, 
bottom color=blue!50!black!20, thick,minimum size=8.1mm,line width=1pt,>=stealth]  

\tikzstyle{contred}=[circle,draw=red,top color=red, 
bottom color=white,thick,minimum size=6mm,line width=1pt,>=stealth]  
\tikzstyle{contyellow}=[circle,draw=yellow,top color=yellow, 
bottom color=yellow!50!black!20,thick,minimum size=6mm,line width=1pt,>=stealth]  
\tikzstyle{contblue}=[circle,draw=blue,top color=blue!80!black!50, 
bottom color=white, thick,minimum size=6mm,line width=1pt,>=stealth]  
\tikzstyle{contgreen}=[circle,draw=green,top color=green!80!black!50, 
bottom color=white, thick,minimum size=6mm,line width=1pt,>=stealth]  
\tikzstyle{contwhite}=[circle,draw=white,color=white, thick,minimum size=6mm,line width=1pt,>=stealth]  
\tikzstyle{contwhiteb}=[circle,draw=white,color=white, thick,minimum size=7.5mm,line width=1pt,>=stealth]  

\tikzstyle{cont}=[circle, draw,
thick,minimum size=6mm,line width=1pt,>=stealth]  

\tikzstyle{contb}=[circle,draw=black!50,top color=white, 
bottom color=darkgreen!50!black!20,thick,inner sep=0pt,minimum size=8.1mm,line width=1pt,>=stealth]  

\tikzstyle{ocont}=[ellipse,draw=blue!50,thick,minimum size=6mm,>=stealth]  
\tikzstyle{blackcont}=[circle,draw=black!50,thick,minimum size=6mm,line width=2pt,>=stealth]  
\tikzstyle{oval}=[ellipse,draw=blue!50,thick,minimum size=6mm,line width=1pt,>=stealth]  
\tikzstyle{ovalb}=[ellipse,draw=blue!50,thick,minimum size=7.5mm,line width=1pt,>=stealth]  
\tikzstyle{disc}=[rectangle,draw=blue!50,thick,line width=1pt,minimum size=6mm]  
\tikzstyle{obs}=[fill=blue!20,thick]  
\tikzstyle{opt}=[star,draw=red!50,thick,minimum size=6mm]  

\tikzstyle{fillred}=[fill=red!20,thick]  
\tikzstyle{fillgreen}=[fill=green!20,thick]  
\tikzstyle{purered}=[fill=red]  
\tikzstyle{state}=[rectangle,fill=red!20]  
\tikzstyle{sobs}=[fill=green!15,thick]  
\tikzstyle{fact}=[fill,minimum size=1.5mm,line width=2pt,>=stealth]
\tikzstyle{varfact}=[draw,minimum size=1.5mm,line width=2pt,>=stealth]
\tikzstyle{sep}=[rectangle,draw=magenta!50,thick,minimum size=6mm]  
\tikzstyle{det}=[fill=red!15,rectangle,draw=red!50,thick,minimum size=6mm]  

\tikzstyle{dethid}=[diamond,draw=red!50,thick,minimum size=6mm]  

\tikzstyle{lineball}=[fill,-*,draw=red!50,line width=1.5pt]
\tikzstyle{redball}=[mark=*,mark options={fill=red!50,draw=red},mark size=0.5pt]
\tikzstyle{greenball}=[mark=*,mark options={fill=green!50,draw=green},mark size=0.5pt]
\tikzstyle{hid}=[circle,draw,thick]  

\tikzstyle{dec}=[rectangle,draw=red!50,thick,minimum size=6mm]  
\tikzstyle{utility}=[diamond,draw=red!50,thick,minimum size=6mm]  
\tikzstyle{contdec}=[circle,draw=blue!50,thick,fill=blue!10,line width=2pt]  
\tikzstyle{decutility}=[diamond,draw=red!50,thick,minimum size=6mm]  

\tikzstyle{contobs}+=[cont]
\tikzstyle{contobs}+=[obs]
\tikzstyle{discobs}+=[disc]
\tikzstyle{discobs}+=[obs]

\tikzstyle{obsred}+=[obs]
\tikzstyle{obsred}+=[red]

\tikzstyle{background grid}=[draw, black!50,step=.1cm]
\tikzstyle{dgraph}=[->, line width=1.5pt]
\tikzstyle{ugraph}=[line width=1.5pt]

\definecolor{magenta}{cmyk}{0.1,1,1,0.5}
\definecolor{darkgreen}{cmyk}{0.6,0.1,0.6,0.6}
\definecolor{pink}{cmyk}{0.1,1,1,0.1}
\definecolor{azzurro}{cmyk}{0.9333, 0.2471, 0.5569, 0.102}
\definecolor{lilla}{rgb}{0.5, 0, 0.5}
\definecolor{mygreen}{rgb}{0, 0.5, 0}
\definecolor{darkorange}{rgb}{1, 0.4, 0} 
\definecolor{darkred}{rgb}{0.8, 0, 0}

\newcommand{\eframe}{\end{frame}}

\newcommand{\bmp}[1]{\begin{minipage}{#1}}
\newcommand{\bmpp}[2]{\begin{minipage}[#1]{#2}}
\newcommand{\emp}{\end{minipage}}
\newcommand{\cb}[1]{\left\{ {#1} \right\}}
\newcommand{\br}[1]{\left( {#1} \right)}
\newcommand{\sq}[1]{\left[ {#1} \right]}

\newcommand{\avp}[2]{\mathbb{E}_{{#1}}\sq{#2}}

\newcommand{\ave}[1]{\mathbb{E}\sq{#1}}

\newcommand{\av}[1]{\left\langle{#1}\right\rangle}

\newcommand{\sett}[1]{\myset{#1}}
\newcommand{\ocm}{\hspace{1cm}}
\newcommand{\hcm}{\hspace{0.25cm}}

\newcommand{\myset}[1]{\mathcal{\uppercase{#1}}}
\newcommand{\beq}{\[}
\newcommand{\eeq}{\]}

\newcommand{\abs}[1]{|#1|}

\tikzstyle{celim}=[circle,draw=red!25,thick,minimum size=6mm,line width=2pt,>=stealth]  %
\tikzstyle{delim}=[draw=red!25]  %

\newcommand{\btz}{\begin{tikzpicture}}
\newcommand{\etz}{\end{tikzpicture}}
\newcommand{\ind}[1]{\mathbb{I}\sq{#1}}
\newcommand{\trans}{^{\textsf{T}}}

\newcommand{\half}{\frac{1}{2}}

\renewcommand{\beq}{\begin{equation}}
\renewcommand{\eeq}{\end{equation}}

\newcommand{\f}[1]{f_{#1}}
\newcommand{\g}[1]{g_{#1}}
\newcommand{\h}[1]{h_{#1}}
\newcommand{\q}[1]{q_{#1}}
\newcommand{\ct}{c^t}
\newcommand{\cp}{c^p}
\newcommand{\vt}{v^t}
\newcommand{\vp}{v^p}

\newcommand{\F}{\mathbb{F}}
\newcommand{\Fp}{\mathbb{F}'}
\newcommand{\Fhat}{{\hat{\mathbb{F}}}}
\newcommand{\FNinf}{{\mathbb{F}^*}}
\newcommand{\FMinf}{{\mathbb{F}'}}
\newcommand{\avc}[1]{\av{#1}}

\renewcommand{\algref}[1]{algorithm(\ref{#1})}
\newcommand{\figref}[1]{fig(\ref{#1})}
\newcommand{\appref}[1]{app(\ref{#1})}
\newcommand{\secref}[1]{sec(\ref{#1})}

\renewcommand{\eqref}[1]{eq(\ref{#1})}

\newcommand{\Falpha}{\mathbb{F}_\alpha}
\newcommand{\FalphaMicro}{\mathbb{F}^\mu_\alpha}

\usepackage[preprint,nonatbib]{neurips_2021}

\usepackage[utf8]{inputenc} 
\usepackage[T1]{fontenc}    
\usepackage{hyperref}       
\usepackage{url}            
\usepackage{booktabs}       
\usepackage{amsfonts}       
\usepackage{nicefrac}       
\usepackage{microtype}      
\usepackage{xcolor}         
\usepackage{authblk}

\title{Sample Efficient Model Evaluation}

\author[1,2]{Emine Yilmaz}
\author[1,2]{Peter Hayes}
\author[1,2]{Raza Habib}
\author[2]{Jordan Burgess}
\author[1,2]{David Barber}
\affil[1]{Department of Computer Science, University College London}
\affil[2]{humanloop.com}
 
\begin{document}

\maketitle

\begin{abstract}
Labelling data is a major practical bottleneck in training and testing classifiers. 
Given a collection of unlabelled data points, we address how to select which subset  to label to best estimate test metrics such as accuracy, $F_1$ score or micro/macro $F_1$. We consider two sampling based approaches, namely the well-known Importance Sampling and we introduce a novel application of Poisson Sampling. For both approaches we derive the minimal error sampling distributions and how to approximate and use them to form estimators and confidence intervals. We show that Poisson Sampling outperforms Importance Sampling both theoretically and experimentally. 
\end{abstract}

\section{Offline Model Evaluation}

How to select {\emph{training}} examples from an unlabelled pool to minimise labelling effort is well studied (see for example \cite{settles2009active}). However, the complementary problem of selecting which testpoints to label to estimate {\emph{test}} performance is less well studied. Having a good estimate of test performance is vital to gain confidence in the predictive performance of a model.  
We focus in this initial work on the `offline' setting in which we assume that a single set of data points will be selected to be labelled. The `online' setting in which labelled testpoints can inform the future selection of testpoints to label is left for a separate study.

We assume that we have a trained probabilistic binary classifier $p(\cp_n|x_n)$, where $x_n$ is an input and $\cp_n\in\cb{0,1}$ is the predicted class, with $\cp_n=1$ being the positive' class and $\cp_n=0$ being the `negative' class. From this trained classifier we assume that a class label is produced deterministically. For example, one may use thresholding to set $\cp_n=1$ if $p(\cp_n=1|x_n)>\theta$ for some user specified $\theta$.  Given a set of test inputs $\sett{X} = \cb{x_1,\ldots, x_N}$ we would like to estimate various measures of performance such as accuracy, $F_1$ score, etc. However, we do not a priori know the true class labels $\ct_n$ and assume that it is very costly (in time/effort) to obtain these true class labels. We therefore know the test set inputs and wish to estimate the test performance using as little test labelling effort as possible. Assuming for the moment that we have access to all true test labels, $\ct_n\in\cb{0,1}$, we first consider test metrics of the form
\beq
\mathbb{F}  \equiv \frac{\sum_{n=1}^N f(\cp_n,\ct_n)}{\sum_{n=1}^N g(\cp_n,\ct_n)}
\label{eq:metric}
\eeq
For example, for the accuracy metric we have $f(\cp,\ct) = \ind{c^{p} = c^{t}}$, $g(\cp,\ct)  =1$. 
Here $\ind{x=y}$ is the indicator function, being 1 when $x=y$ and 0 otherwise.  Similarly, for the $\Falpha$ metric $0\leq\alpha\leq 1$
\beq
\Falpha \equiv  \frac{\sum_{n=1}^N \ind{c^{t}_n=1}\ind{c^{p}_n=1}}{\alpha \sum_{n=1}^N \ind{c^{p}_n=1} + (1-\alpha)\sum_{n=1}^N \ind{c^{t}_n=1}}  
\eeq
where
\begin{align}
f(\cp,\ct) &= \ind{c^{t}=1}\ind{c^{p}=1},\hcm g(\cp,\ct) = \alpha\ind{c^{p}=1}+(1-\alpha)\ind{\ct=1}.
\end{align}
The $F_1$ metric is given by $\mathbb{F}_\half$; recall ($\F_0$) and precision ($\F_1$) along with other metrics are also easily defined. Where there is no ambiguity, we write $\f{n}$ in place of $f(\cp_n,\ct_n)$, and similarly for $g_n$.



The above metrics require us to know the true value of the test label $\ct_n$. However, in our scenario we  wish to get a good estimate of the test metric, without having to label all test points. A simple approach is to uniformly sample a subset of test points $x_n$, label them and calculate the performance.  However, this is sub-optimal, particularly in the case of high class imbalance which occurs frequently in practice. Consider for example the ability to correctly classify an offensive tweet for content moderation. Only a small percentage of the dataset (say less than 7\%) may contain offensive tweets\footnote{See for example \href{https://www.kaggle.com/vkrahul/twitter-hate-speech}{www.kaggle.com/vkrahul/twitter-hate-speech}}. 
We therefore wish to actively select datapoints to label such that we obtain an accurate metric (for example, by concentrating on tweets which are likely to be offensive). 

We consider two sampling-based approaches to estimating test metrics --  Importance Sampling and  Poisson Sampling. Full derivations are in the appendix, including the more general approach in \appref{sec:multi} that can be used to estimate for example the macro $F_1$ score in multi-label classification. The theory behind the approaches is similar and we begin with the better known Importance Sampling. We note that an ideal estimator would have the property that as all $N$ test labels are known, the metric will be correctly evaluated with no uncertainty (for a deterministic true classifier $p(\ct_n|x_n)$). 
 
It is important to bear in mind the two stages in which we need estimates of the metric. The  pre-sampling stage estimates the metric to approximate the optimal sampler. The post-sampling stage uses the drawn samples to form an estimate of the performance. 
We also note that the metric used to form the samples may not be the same as the metric used for the evaluation. This is because we will typically only wish to use a single criterion to determine which test points should be labelled -- however, we may wish to estimate a variety of performance metrics using those samples. For example, we may decide which datapoints to label on the basis of getting the best estimate of $F_1$ score, but also use those samples to estimate $F_1$,  recall, precision, etc. 
 
\section{Importance Sampling\label{sec:is}}

We first extend the work of \cite{activeF} to a more general class of test metrics.  Importance Sampling (IS) can be used to estimate test performance by independently sampling (with replacement)  indices $m_1,\ldots, m_M$,  $m_i\in\cb{1,\ldots, N}$ from the test dataset using the importance distribution $\q{n}$, $n\in\cb{1,\ldots, N}$.  We then form the estimator using
\beq
\hat{\mathbb{F}} = \frac{\hat{x}}{\hat{y}}
\label{eq:Fhat}
\eeq
where $\hat{x}$ and $\hat{y}$ are obtained from $M$ sampled datapoints:
\beq
\hat{x}=\frac{1}{MN}\sum_{i=1}^M \frac{\f{m_i}}{\q{m_i}}, \ocm \hat{y}=\frac{1}{MN}\sum_{i=1}^M \frac{\g{m_i}}{\q{m_i}}.
\label{eq:xyhat}
\eeq

%
Both $\hat{x}$ and $\hat{y}$ are sums of independently distributed random variables. For large $M$ (and any value of $N$), $(\hat{x},\hat{y})$ will therefore be approximately jointly Gaussian distributed (per the Central Limit Theorem, \cite{Stats}).  It is straightforward to show (\appref{app:IS}) the Gaussian $p(\hat{x},\hat{y})$ has mean
\beq
\mu_x \equiv \ave{\hat{x}} =\frac{1}{N}\sum_{n=1}^N \avc{\f{n}}, \ocm 
\mu_y \equiv \ave{\hat{y}} = \frac{1}{N}\sum_{n=1}^N \avc{\g{n}}.
\eeq
Here $\avc{\cdot}$ is the expectation with respect to the true label distribution $p(\ct_n|x_n)$. The corresponding covariance elements are given by
\beq
\Sigma_{xy} = \frac{1}{M} \br{\frac{1}{N^2}\sum_{n=1}^N \frac{\avc{\f{n}\g{n}}}{\q{n}} - \mu_x\mu_y},
\eeq
and similarly for $\Sigma_{xx},\Sigma_{yy}$.  Since $\q{n}$ is typically $O(1/N)$, the mean elements are $O(1)$ whilst the covariance elements scale as $O(1/M)$ meaning that for large $M$ fluctuations from the mean will typically be small.  Writing $\hat{x}$ and $\hat{y}$ in terms of a mean-fluctuation decomposition, and expanding for a small fluctuation $\Delta$, 
\beq
\hat{\mathbb{F}} = \frac{\hat{x}}{\hat{y}} = \frac{\mu_x + \Delta_x}{\mu_y + \Delta_y}= \frac{1}{\mu_y}\frac{\mu_x + \Delta_x}{1 + \frac{\Delta_y}{\mu_y}}\approx \frac{1}{\mu_y}\br{\mu_x + \Delta_x}\br{1-\frac{\Delta_y}{\mu_y}}
\label{eq:delta:rep},
\eeq
hence
\beq
\ave{\hat{\mathbb{F}}} = \frac{\mu_x}{\mu_y}\br{1 - \frac{\Sigma_{xy}}{\mu_x\mu_y}} + O(1/M^2).
\label{eq:av:f}
\eeq
The expected metric therefore tends to
\beq
\FMinf \equiv \frac{\mu_x}{\mu_y} = \frac{\sum_{n=1}^N \avc{f_n}}{\sum_{n=1}^N \avc{g_n}}
\label{eq:FMinf}
\eeq
as the number of IS samples $M\rightarrow\infty$. This is the exact expected value of the metric calculated on the test set. The estimator $\Fhat$ is therefore a consistent estimator of $\Fp$, with bias  $O(1/M)$.

\subsection{Optimal Sampling Distribution}

The squared error between the finite $M$ estimator $\Fhat$ and infinite $M$ limit $\FMinf$
is (to leading order in $1/M$ - see \appref{sec:opt:is:error} for more details)
\beq
\ave{\br{\Fhat  - \FMinf}^2}= \frac{1}{MN^2\mu_y^2}\sum_{n=1}^N \frac{h_n^2}{q_n},
\label{eq:is:error}
\eeq
where $h_n^2\equiv \avc{\br{f_n - \FMinf g_n}^2}$ is given by
\begin{multline}
h_n^2 = p(c_n^t=1|x_n) \br{f(\cp_n,\ct_n=1) -\FMinf g(\cp_n,\ct_n=1)}^2\\
+p(\ct_n=0|x_n) \br{f(\cp_n,\ct_n=0) -\FMinf g(\cp_n,\ct_n=0)}^2.
\label{eq:h:def}
\end{multline}
The minimal error estimator is then given by $q_n\propto \abs{h_n}$ (see \appref{sec:opt:is}).
To calculate the optimal sampler, we therefore need to know the true class distribution $p(\ct_n=1|x_n)$. In related works (eg \cite{Bach-2007,activeF}) the assumption $p(\ct_n=1|x_n) = p(\cp_n=1|x_n)$ is made. However, this places great faith in the model and we found this can lead to overconfidence, particularly in models with very high or very low probabilities. To address this we replace the unknown $p(\ct_n|x_n)$ with an estimate 
\beq
p_a(\ct_n=1|x_n) = \lambda p(\cp_n=1|x_n) + (1-\lambda)0.5
\label{eq:pred:approx}
\eeq
for some user chosen $0\leq\lambda\leq 1$. In our experiments we found that $\lambda=0.9$ works well in practice.\footnote{Any choice of $\lambda$ gives in a consistent estimator, but a good choice of $\lambda$ can result in better convergence.} 
We also use this to form a pre-sampling  approximation to $\FMinf$ by computing the expectation with respect to $p_a(\ct_n|x_n)$. We denote this
\beq
\Fp_a\equiv \frac{\sum_{n=1}^N \avp{a}{f_n}}{\sum_{n=1}^N \avp{a}{g_n}}.
\label{eq:Fa}
\eeq
which is used in place of $\Fp$ to evaluate \eqref{eq:h:def}. This enables us to fully define $h_n$ and the optimal Importance Sampler $q_n\propto \abs{h_n}$. 

\begin{algorithm}[t]
	\caption{Optimal Importance Sampling\label{alg:opt:is}} 
	\begin{algorithmic}[1]
	\State {\bf Define Sampler and Draw Samples}:
	\State Choose the metric $f,g$ to define the sampling distribution
	\State Choose $\lambda$ and define the predicted class distribution $p_a(\ct_n|x_n)$ using \eqref{eq:pred:approx}
	\State Calculate the predicted metric $\Fp_a$ using \eqref{eq:Fa} and the squared deviation $h_n^2$ defined in \eqref{eq:h:def} using $\Fp_a$ in place of the true $\Fp$ and $p_a(\ct_n|x_n)$ in place of the true  $p(\ct_n|x_n)$
	\State Define the Importance distribution $q_n\propto \abs{h_n}$ and draw $M$ samples (with replacement) using $q$
	\State {\bf Post Sampling Metric Approximation}:
\State Define the evaluation metric to be used $f,g$
	\State Calculate the metric $\Fhat$ defined in \eqref{eq:Fhat} using the drawn samples and the evaluation metric
	\State Calculate the sampled value  $\hat{y}$ using \eqref{eq:post:g} and  error $\sigma^2$ using \eqref{eq:is:error:post}
	\State Calculate confidence limits assuming a Beta distribution with mean $\Fhat$ and variance $\sigma^2$
	\end{algorithmic} 
\end{algorithm}

\subsection{Post-sampling Performance Approximation}


Given the samples, and the sampling distribution $q$ from which they were drawn, we estimate the metric using \eqref{eq:Fhat} for the given choice of test metric to form a post-sample estimate $\Fhat$. For an estimate of the error of this metric estimate, we can again use IS
\beq
\frac{1}{N}\sum_{n=1}^N \frac{\h{n}^2}{\q{n}}\approx \frac{1}{MN}\sum_{i=1}^M \frac{\h{m_i}^2}{\q{m_i}^2},
\label{eq:post:sq:dev}
\eeq
in which the $\h{m_i}$ are evaluated at the sampled true labels.\footnote{In extreme cases (no true positive cases are drawn in the testset) this can result in a value of $h_n=0$, giving an overconfident estimation. 
%
%
For this reason, we add a small value $\epsilon=1\times e^{-10}$ to each $\h{n}^2$ in \eqref{eq:post:sq:dev}.} 
Similarly, $\Fp$ is replaced with our post-sample estimate $\Fhat$ to evaluate $\h{m_i}$. We also use the post-sample estimate
\beq
\mu_y \approx \frac{1}{MN}\sum_{i=1}^M \frac{\g{m_i}}{\q{m_i}}\equiv \hat{y}.
\label{eq:post:g}
\eeq
Using these we obtain a post sampling error approximation of
\beq
\sigma^2 \equiv \ave{\br{\Fhat  - \FMinf}^2}
\approx
\frac{1}{\hat{y}^2}\frac{1}{(MN)^2}\sum_{i=1}^M \frac{\h{m_i}^2}{\q{m_i}^2}.
\label{eq:is:error:post}
\eeq
The full procedure is given in \algref{alg:opt:is}.
To form sensible confidence limits around the metric estimator, we fit a Beta distribution with mean $\Fhat$ and variance $\sigma^2$.

\section{Poisson Sampling}

Importance Sampling has some clear drawbacks:  For a deterministic true classifier, once a datapoint has been sampled we know its true label. The IS estimator only becomes exact in the limit of infinite number of samples $M$ (and all datapoints are guaranteed to be touched). More seriously, in IS the requirement $q_n>0$, $\sum_n q_n=1$ means that no datapoint can have $q_n=1$. Hence, even for datapoints which would cause large errors if not included, they cannot be selected with certainty. On the other hand, Poisson Sampling (see eg \cite{SurveySampling,CSS}) has the property that as the number of samples tends to the number of datapoints, the approximation becomes exact. As we will show, it also has the ability to select with certainty datapoints that should be labelled.


We again consider the problem of approximating $\F = \sum_{n=1}^N f_n/\sum_{n=1}^N g_n$. The Poisson Sampler  can be used to form an estimator
\beq
\Fhat = \frac{\hat{x}}{\hat{y}}, \ocm \hat{x} = \frac{1}{N}\sum_{n=1}^N \frac{s_n}{b_n}f_n , \hcm \hat{y} = \frac{1}{N}\sum_{n=1}^N \frac{s_n}{b_n}g_n.
\label{eq:fhat:bs}
\eeq
Here the binary indicators ${s_n\in\cb{0,1}}$ are independently drawn from a set of Bernoulli distributions $p(s_n=1)=b_n$, $b_n\in\sq{0,1}$. We also refer to this as Bernoulli Sampling\footnote{In the survey sampling literature, ``Bernoulli Sampling" refers to Poisson Sampling in which the Bernoulli probabilities of including each sample are all equal. Here we will use the term Bernoulli Sampling interchangeably with Poisson Sampling (with different inclusion probabilities).}.

The $s_n$ select which test datapoints need to be labelled since
\beq
\frac{1}{N}\sum_{n=1}^N \frac{s_n}{b_n}f_n = \frac{1}{N}\sum_{n: s_n=1}\frac{f_n}{b_n}.
\eeq
Setting all $b_n=1$ (which means that all test points would be drawn with probability 1), we have $\Fhat= \sum_{n=1}^N f_n/\sum_{n=1}^N g_n$ which, for a deterministic true classifier, is equal to the exact value $\Fp$ of the metric on the test data. Hence, as all $b_n\rightarrow 1$, the estimator is consistent. 

Since $\hat{x}$ and $\hat{y}$ are sums of independently generated random variables, for large $N$, $p(\hat{x},\hat{y})$ will be approximately Gaussian distributed \cite{Stats} with mean 
\beq
\mu_x = \frac{1}{N}\sum_n \frac{\ave{s_n f_n}}{b_n}=\frac{1}{N}\sum_n \avc{f_n}, \ocm \mu_y = \frac{1}{N}\sum_n \avc{g_n},
\eeq
where $\ave{s_n f_n}=\ave{s_n}\av{f_n}=b_n\av{f_n}$. Here we used the fact that for a 0/1 binary variable $\ave{s_n}=p(s_n=1|b_n)=b_n$. The covariance elements are also straightforward to calculate:
\beq
\Sigma_{xy} = \frac{1}{N^2}\sum_{n=1}^N \br{\frac{\avc{f_ng_n}}{b_n}-\avc{f_n}\avc{g_n}}
\eeq
and similarly for $\Sigma_{xx},\Sigma_{yy}$.  Since the covariance elements are $O(1/N)$ compared to the $O(1)$ mean elements,  fluctuations $\Delta$ from the mean are typically small and we can write
\beq
\hat{\mathbb{F}} = \frac{\hat{x}}{\hat{y}} = \frac{\mu_x + \Delta_x}{\mu_y + \Delta_y}= \frac{1}{\mu_y}\frac{\mu_x + \Delta_x}{1 + \frac{\Delta_y}{\mu_y}}\approx \frac{1}{\mu_y}\br{\mu_x + \Delta_x}\br{1-\frac{\Delta_y}{\mu_y}},
\label{eq:delta:rep:two}
\eeq
hence
\beq
\ave{\hat{\mathbb{F}}} = \frac{\mu_x}{\mu_y}\br{1 - \frac{\Sigma_{xy}}{\mu_x\mu_y}} + O(1/N^2)= \frac{\mu_x}{\mu_y} + O(1/N).
\eeq
In the deterministic true classifier setting, the bias of this estimator is therefore approximately
\beq
\frac{1}{N^2\mu_y^2}\sum_{n=1}^N{f_ng_n}\br{\frac{1}{b_n}-1},
\label{eq:bias:det}
\eeq
which scales as $O((1/b-1)/N)$ for typical values $b$ of $b_n$.  The expected error of the estimator in approximating the metric is (to leading order in $1/N$), see \appref{app:bs}
\begin{align}
\ave{\br{\Fhat  - \Fp}^2} &\approx\frac{1}{N^2\mu_y^2}\sum_{n=1}^N\br{\frac{1}{b_n}\avc{\br{f_n - \Fp g_n}^2} - \br{\avc{f_n}-\Fp\avc{g_n}}^2}.
\label{eq:var:est:two}
\end{align}
In the deterministic true classifier setting, this reduces to 
\beq
\ave{\br{\Fhat  - \Fp}^2} \approx \frac{1}{N^2\mu_y^2}\sum_{n=1}^N{\br{f_n - \Fp g_n}^2}\br{\frac{1}{b_n}-1}.
\label{eq:var:est:det}
\eeq

\begin{algorithm}[t]
	\caption{Optimal Bernoulli Sampling\label{alg:bs}} 
	\begin{algorithmic}[1]
	\State {\bf Define Sampler and Draw Samples}:
	\State Choose the metric $f,g$ to define the sampling distribution
	\State Choose $\lambda$ and define the predicted class distribution $p_a(\ct_n|x_n)$ using \eqref{eq:pred:approx}
	%
	\State Calculate the predicted metric $\Fp_a$ using \eqref{eq:Fa} and the squared deviation $h_n^2$ defined in \eqref{eq:h:def} using $\Fp_a$ in place of the true $\Fp$ and $p_a(\ct_n|x_n)$ in place of the true  $p(\ct_n|x_n)$
	\State Choose $M$, get the weights $b_n$ using the algorithm in \secref{sec:BS:optb} and draw samples $s_n$
	\State {\bf Post Sampling Metric Approximation}:
\State Define the evaluation metric to be used $f,g$
	\State Calculate the metric $\Fhat$, \eqref{eq:fhat:bs} using the drawn samples and the evaluation metric
	\State Calculate the error $\sigma^2$ using  \eqref{eq:var:est:stoch} and $\Fhat$ in place of $\Fp$
	\State Calculate confidence limits assuming a Beta distribution with mean $\Fhat$ and variance $\sigma^2$
	\end{algorithmic} 
\end{algorithm}

\subsection{Optimal Sampling Distribution}

The $b$-dependence of the error in \eqref{eq:var:est:two} is given by $\sum_n h_n^2/b_n$ where $h_n^2 =  \avc{\br{f_n - \Fp g_n}^2}$; $h_n$ can be estimated taking the expectation with respect to an approximation $p_a(\ct_n|x_n)$, leading to the same definition \eqref{eq:h:def}.
Clearly, we can minimise the error to zero by setting all $b_n=1$. However, that means that we would simply sample all test points. We therefore add the constraint $\sum_{n=1}^N b_n=M$ so that the expected number of test points that will be sampled is $M$;  
%
%
the variance of the number of sampled points is $M - \sum_{n=1}^N b_n^2$.  The objective  $\sum_n h_n^2/b_n$ is convex on the feasible set $0\leq b_n\leq 1$, $\sum_{n=1}^N b_n=M$ and an efficient $O(N)$ algorithm to find the global minimum is given in \appref{sec:BS:optb}, similar to a water-filling algorithm, see for example \cite{Qi2012}.

\subsection{Post-sampling Performance Approximation}


We estimate the error by sampling, using
\begin{align}
\ave{\br{\Fhat - \Fp}^2} &\approx\frac{1}{N^2\mu_y^2}\sum_{n=1}^N\frac{s_n}{b_n}\br{\frac{1}{b_n}\av{\br{f_n - \Fp g_n}^2} -\br{\av{f_n} - \Fp \av{g_n}}^2}.
\label{eq:var:est:stoch}
\end{align}
For a deterministic true classifier the above becomes
\beq
\ave{\br{\Fhat - \Fp}^2} \approx\frac{1}{N^2\mu_y^2}\sum_{n=1}^N\frac{s_n}{b_n} h_n^2,
\eeq
where $h_n^2  = \br{\frac{1}{b_n}-1}\br{f_n - \Fp g_n}^2$. As for IS, to avoid issues when there are no true positives, we add a small value $\epsilon$ to each $h^2_n$. 
We approximate the denominator using
\beq
N\mu_y \approx \sum_{n=1}^N \frac{s_n}{b_n}g_n
\label{eq:post:error:den}.
\eeq
The full procedure is given in \algref{alg:bs}. In contrast to IS, there are some key differences:

\begin{itemize}[leftmargin=*]



\item 
In BS, for a deterministic true classifier $p(\ct_n|x_n)$ and $M=N$ samples, we will evaluate $f_n$, $g_n$ exactly at each point, resulting in calculating the exact metric.  In contrast, in IS, as ${M\rightarrow\infty}$ the estimator becomes exact (for any testset size $N$). 

\item In BS datapoints can be included with probability 1.  In \figref{fig:weights} we show the probability of datapoint inclusion $b_n$ for the Bernoulli approach and the IS inclusion probabilities (the probability that a datapoint $n$ will be included in any of the $M$ importance samples) $\pi_n =  1 - (1-q_n)^M$.  

\item The BS weights start to saturate to 1 when $h^* \geq \frac{N}{M}\bar{h}$ where $h^*$ is the maximum of the deviations $h_1,\ldots, h_N$ and $\bar{h}$ is their average, see \appref{sec:is:vs:bs}.

\item In BS we do not know exactly how many datapoints will be sampled, since this is stochastic. In IS we know the number of samples drawn $M$, but the number of unique samples is stochastic.
\end{itemize}
An important question is whether the asymptotic performance of BS is superior (in expectation) to IS. In \appref{sec:is:vs:bs} we consider the error estimates and show that BS indeed has a lower expected error than IS. This theoretical result is borne out by our experiments, \secref{sec:exp}.

\section{Related Work\label{sec:related}}

In \cite{activeF} the authors consider F-metrics and derive the optimal Importance distribution (to minimise estimator variance) under the assumption of asymptotic normality.  In \secref{sec:is} we generalised their results to metrics of the form $\sum_n{f_n}/\sum_n{g_n}$ -- further generalisation, for example to macro $F_1$ is given in \secref{sec:multi}. More importantly, we derived a new approach based on Bernoulli Sampling, which we showed numerically and theoretically is superior to Importance Sampling.

%
%
%

In \cite{Welinder} the authors model the relationship between a scalar score $s(x)$ for a model $p(\cp|s(x))$ and the true label $\ct$. 
How to select the label pairs $(s(x),\ct)$ to learn this relationship is not stated and, in cases of severe class imbalance, selecting these uniformly at random is likely to be suboptimal. 
%
Another approach is to use stratified sampling \cite{Druck} in which the testset is split into a number of defined strata (for example based on the probability $p(\cp=1|x_n)$). In \cite{Druck} the authors show that this can reduce the variance of estimating $\sum_n f_n$. In our case, the metrics of interest $\sum_n {f_n}/\sum_n {g_n}$ are not of the form to which \cite{Druck} can be applied. Implementing  stratified sampling using a perturbation approximation of the ratio $\sum_n f_n/\sum_n g_n$ would make for an interesting but separate study to ours.

Stratified sampling and importance sampling techniques have also been applied to information retrieval metrics such as average precision and normalised discounted cumulative gain \cite{Yilmaz08Stratified, pavlu07practical, Aslam06Importance}. However, these works  focused on deriving unbiased estimations of evaluation metrics \emph{given} a sampling distribution, as opposed to identifying the minimum error sampling distribution.

In \cite{MarchantRubinstein} the authors recognise that in \cite{activeF} the optimal Importance distribution depends on unknown quantities (such as an estimate of the true $F$ measure). They update `online' the estimate of the $F$ metric as samples are obtained and subsequently update the sampling distribution. This results in an iterative approach to continually updating the sampler.

\begin{figure}
\centering
\includegraphics[width=0.31\textwidth]{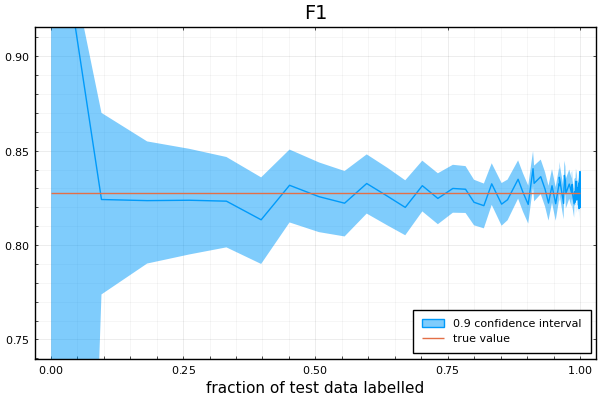}\includegraphics[width=0.3\textwidth]{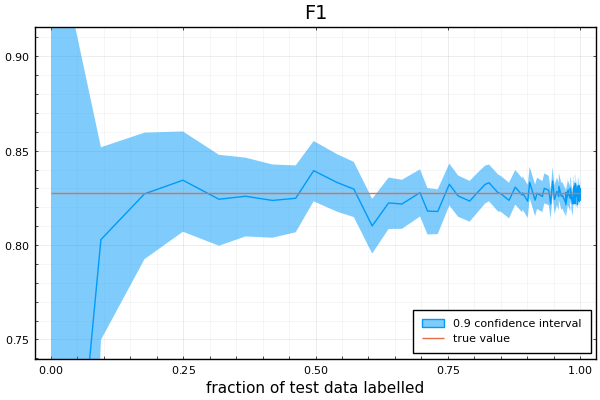}\includegraphics[width=0.3\textwidth]{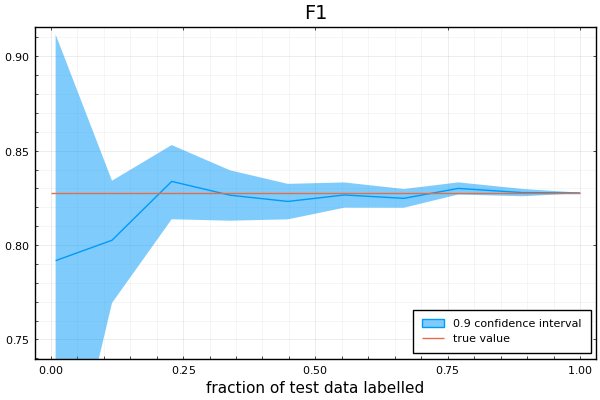}
\caption{Estimating the $F_1$ Test Metric for MNIST. Left: Uniform Sampling. Middle: Importance Sampling. The x-axis is the number of distinct samples as a fraction of the test data (not the total number of samples drawn including repetitions). Right: Bernoulli Sampling. See also \secref{sec:add:results}.\label{fig:mnist:is:bs}}
\end{figure}



In \cite{pmlr-v80-nguyen18d} the test data is split into labelled and unlabelled sets, with a model fitted to predict the label of the remaining unlabelled data. 
The model is updated online to decide which points to label as new data is labelled. Whilst there is empirical support for their approach, there seems no clear theoretical support.  In \cite{kossen2021active} (similar to \cite{MarchantRubinstein,pmlr-v80-nguyen18d}) the Importance sampler is updated online as test labels are observed; the method is however limited to simple metrics of the form $\sum_n f_n$.

Unlike IS, in which a data index $n$ can be selected one at a time, BS selects a \emph{collection} of indices. To dispel potential criticism of BS, we note that it is still possible to do online BS in which at each round a collection of the remaining unsampled indices are drawn, correcting for any introduced bias, see \appref{app:online}.  Our focus here though is the offline setting and we leave the online setting to a separate study. 



\section{Experiments\label{sec:exp}}

We compare Uniform (IS with a uniform distribution), Importance and Bernoulli Sampling with additional results in \secref{sec:add:results}. Code is available\footnote{\url{http://www.filedropper.com/testestimationpublic} MIT License} -- all experiments used a Ryzen 3400G.

\subsection{MNIST\label{sec:mnist}}

A 3 layer Neural Network binary classifier is trained on MNIST \cite{lecun-mnisthandwrittendigit-2010} to predict whether an image is an "8" or not. This is unbalanced problem with 9 times as many "negative" examples (non "8s") to positive examples ("8s"). The Bernoulli and Importance Samplers are chosen to minimise the $F_1$ metric since this is sensitive to class imbalance and a reasonable overall performance metric. In practice, users will typically wish to evaluate a variety of metrics given the sampled test labels.  To mimic this we use the resulting samples to estimate Accuracy, $F_1$, Precision, Recall and Sensitivity. The estimators (and 90\% confidence limits) are shown in figs(\ref{fig:mnist:is:bs},\ref{fig:mnist:is:bs:app}) for a single experiment whilst increasing the number of examples that are used to estimate the metrics. The total number of examples available in the testset is $N=11200$, split equally between the 10 classes. For this toy problem we can calculate the true metric and therefore evaluate how accurate the estimates are. 
For this problem, the true classifier is deterministic (we have a single label for each test datapoint) and the Bernoulli Sampler therefore returns the exact metric as $M$ increases towards $N$. 

\begin{figure}
\centering
\includegraphics[width=0.31\textwidth]{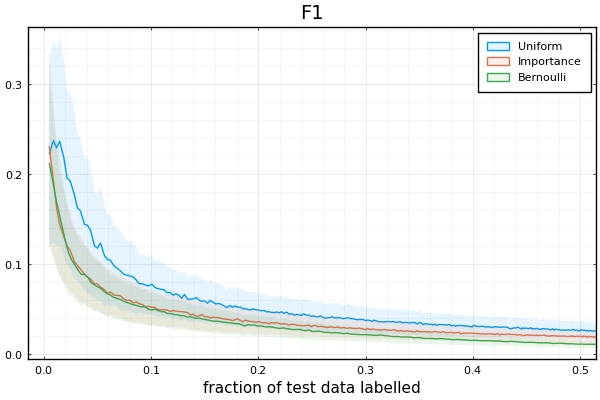}
\includegraphics[width=0.31\textwidth]{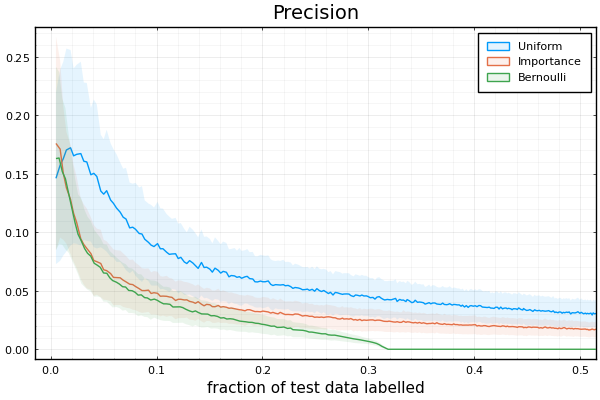}
\includegraphics[width=0.31\textwidth]{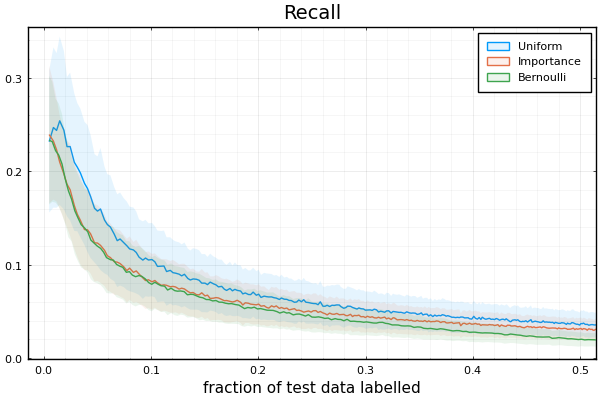}
\caption{Estimating Test Metrics for MNIST. The y-axis is the expected error in each metric: $ |\text{true}-\text{estimate}|$ averaged over 3000 experiments, along with 0.5 standard deviation.  The x-axis is the number of distinct samples as a fraction of the test data. 
%
\label{fig:mnist:new:us:is:bs}}
\end{figure}

\begin{figure}[t]
\centering
\includegraphics[width=0.32\textwidth]{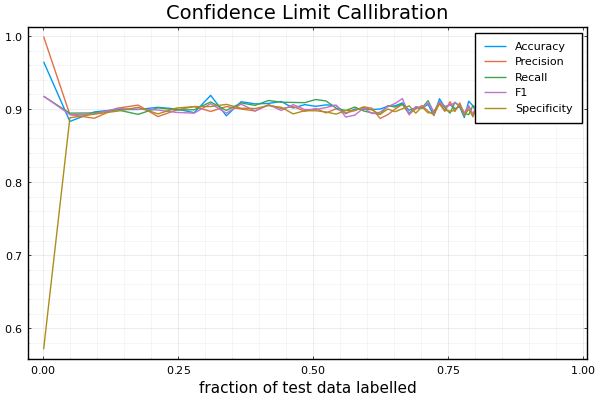}
\includegraphics[width=0.32\textwidth]{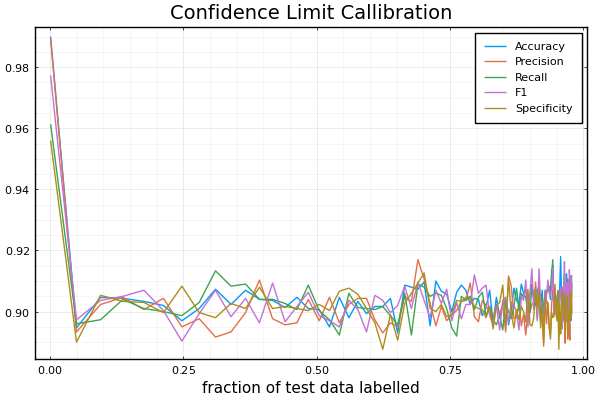}
\includegraphics[width=0.32\textwidth]{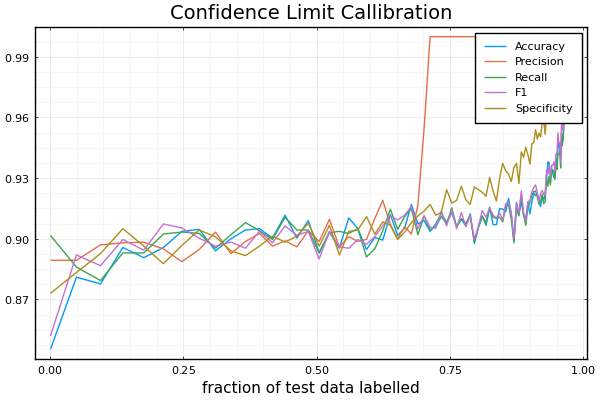}
\caption{Calibrating the confidence limits for MNIST. We ran 3000 experiments with a desired confidence limit of 90\% meaning that we would expect 90\% of the time that the true metric lied within the confidence limit. Left: Uniform Sampling. Middle: Importance Sampling. Right: Bernoulli Sampling. The x-axis is the number of distinct samples as a fraction of the test data; the y-axis is the fraction of the experiments for which the true metric lied within the predicted confidence limit. All methods have roughly the correct behaviour, meaning that the confidence limits are well calibrated. For Bernoulli Sampling, there is deviation as the fraction of the dataset labelled tends to 1 and the metrics become exactly calculated. 
\label{fig:mnist:call:us:is:bs}}
\end{figure}

By repeating the experiments we can calculate the error in the estimates, see figs(\ref{fig:mnist:new:us:is:bs},\ref{fig:mnist:log:us:is:bs:app}), with Bernoulli Sampling significantly outperforming Importance Sampling and Uniform Sampling as the number of labelled datapoints increases. The uncertainty estimates are reasonably well calibrated, see \figref{fig:mnist:call:us:is:bs} for all three methods. This can be quite effective, even when the size of the sampled dataset is small; however, in extreme limits (of a very small number of samples $M$) or when the estimator becomes exact (in the Bernoulli setting) then the predicted confidence limits become less accurate.  In \appref{sec:cross:results} we show (for both IS and BS) that if we wish to get the best post-sampling estimate of metric $X$, there is no better pre-sampling metric to use than $X$.

\begin{figure}
\centering
\includegraphics[width=0.32\textwidth]{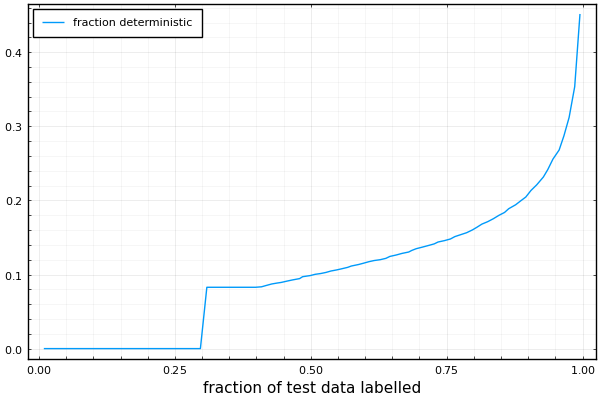}
\includegraphics[width=0.32\textwidth]{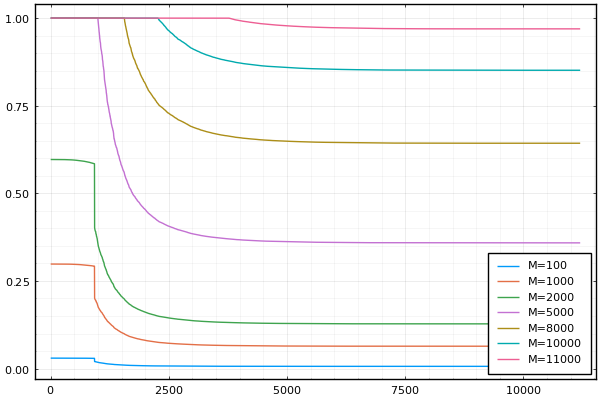}
\includegraphics[width=0.32\textwidth]{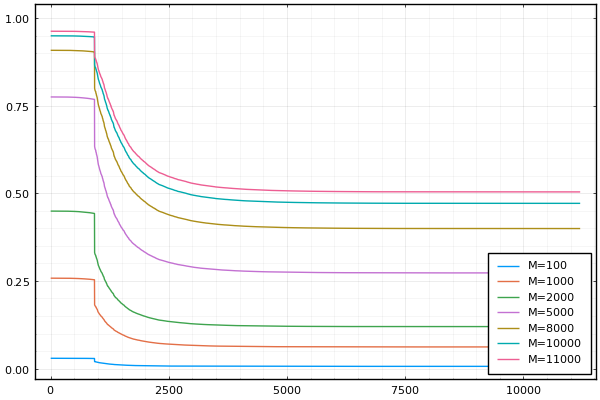}
\caption{Estimating Test Metrics for MNIST based on optimising for $F_1$ score. Left: the y-axis is the fraction of the testpoints that are selected deterministically (those that have $b_n=1$). As the fraction of data labelled increases, points become increasingly deterministically selected. Middle: Bernoulli weights (ordered highest first) for the MNIST classification based on optimising for $F_1$ score. The y-axis is the value $b_n$ for each of the $N=11200$ datapoints. As the number of labelled points $M$ increases, the datapoints become increasingly deterministically selected. Right: Importance Sampling inclusion weights $\pi_n$, ordered highest first. Even though potentially problematic datapoints have relatively high weight they can never be selected with certainty. \label{fig:weights}}
\end{figure}

The Bernoulli Sampler has the interesting property that test datapoints can be selected with probability 1. In \figref{fig:weights} we show the fraction of samples that are drawn with probability 1 ($b_n=1$). For relatively small $M$, no datapoints are drawn deterministically; there is a transition to a finite fraction drawn deterministically after around 25\% of the dataset has been labelled.

\subsection{20 Newsgroups}

We used the scikitlearn\footnote{\tiny{\url{https://scikit-learn.org/stable/tutorial/text_analytics/working_with_text_data.html}}} dataset which contains approximately 20,000 newsgroup documents, partitioned (nearly) evenly across 20 different newsgroups.\footnote{\tiny{\url{http://qwone.com/~jason/20Newsgroups/}}} We used a simple Naive Bayes classifier, based on the scikitlearn train (11314 datapoints) test set (11313 datapoints) split. As for MNIST, we draw samples according to the $F_1$ metric and evaluate the log error for a variety of metrics as we increase the number of test datapoints labelled. The results in \figref{fig:mnist:log:us:is:bs:ng:app} show that Bernoulli Sampling is on average superior to Optimal Importance and Uniform Sampling.

\subsection{Toxic Comment Classification Challenge\label{sec:toxic}}

\begin{wrapfigure}[14]{r}{0.45\textwidth}
\vspace{-7.5mm}
\begin{center}
\includegraphics[width=0.45\textwidth]{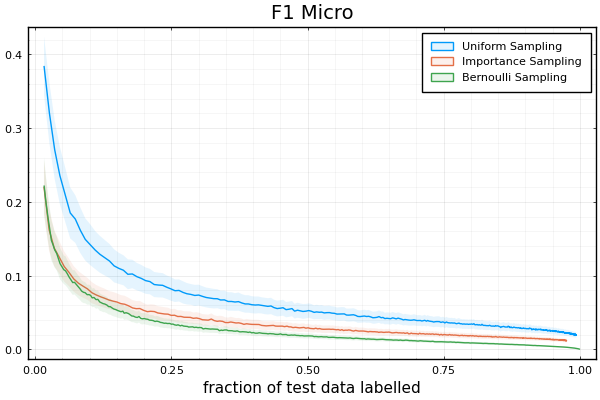}
\end{center}
\caption{Toxic Comment absolute error in estimating the micro $F_1$. Average over 3000 runs with 0.25 standard deviations shown.} 
\label{fig:jigsaw:us:is:bs}
\end{wrapfigure}
This is a multi-label Kaggle Challenge\footnote{\tiny{\url{https://www.kaggle.com/c/jigsaw-toxic-comment-classification-challenge}}} to classify tweets. There are 6 binary classes: toxic, severe toxic, obscene, threat, insult, identity hate.  We used a Naive Bayes model\footnote{\tiny{\url{https://towardsdatascience.com/journey-to-the-center-of-multi-label-classification-384c40229bff}}} for each of the 6 binary classifiers, based on simple features extracted from tweets.  How to extend our approach to both micro and macro $F_1$ metrics is explained in \secref{sec:multi}, along with results, figs(\ref{fig:jigsaw:us:is:bs}, \ref{fig:multi:bs}).  This is a challenging problem since in the 600 test points, there are only 125 non-zero positive values out of a possible $600\times 6 = 3600$ positive values. In this case the larger number of classes means that the potential deviations $h_n$ are large and BS starts to significantly outperform IS for small fractions of the test data being labelled. 


\section{Summary\label{sec:summary}}

We addressed the important challenge of approximating the test performance for common metrics used in single and multi-label classification problems. This work addressed the `offline'  scenario in which test points are selected {\emph{without}} the new labels being able to inform future points to label. 
We introduced Bernoulli Sampling and compared it to the more standard Importance Sampling to estimate these performance measures.

Our conclusion is that whilst Optimal Importance Sampling and Bernoulli Sampling perform the same in the limit of very small fractions of labelled data, Bernoulli Sampling inevitably outperforms Importance Sampling as the amount of labelled data increases.  

Bernoulli Sampling has better accuracy and a lower variance in the examples we considered and we believe it is generally superior for estimating test performance since potentially problematic test points can be included in the sample with certainty.  Bernoulli Sampling (Poisson Sampling) is straightforward to implement and, given its superiority over Importance Sampling, we see it as a drop-in replacement for Importance Sampling for performance estimation.  

\newpage

\bibliographystyle{unsrt}

\begin{thebibliography}{10}

\bibitem{settles2009active}
B.~Settles.
\newblock {Active Learning Literature Survey}.
\newblock Computer Sciences Technical Report 1648, University of
  Wisconsin--Madison, 2009.

\bibitem{activeF}
C.~Sawade, N.~Landwehr, and T.~Scheffer.
\newblock {Active Estimation of F-Measures}.
\newblock In {\em Advances in Neural Information Processing Systems 23}, pages
  2083--2091, 2010.

\bibitem{Stats}
Y.~Dodge.
\newblock {\em {The Concise Encyclopedia of Statistics}}.
\newblock {Springer}, 2008.

\bibitem{Bach-2007}
F.~Bach.
\newblock Active learning for misspecified generalized linear models.
\newblock In B.~Sch\"{o}lkopf, J.~Platt, and T.~Hoffman, editors, {\em Advances
  in Neural Information Processing Systems}, volume~19. MIT Press, 2007.

\bibitem{SurveySampling}
C-E. S\"arndal, B.~Swensson, and J.~Wretman.
\newblock {\em {Model Assisted Survey Sampling}}.
\newblock {Springer}, 1992.

\bibitem{CSS}
A.~Botev, B.~Zheng, and D.~Barber.
\newblock {Complementary Sum Sampling for Likelihood Approximation in Large
  Scale Classification}.
\newblock In {\em {AISTATS}}, volume~54 of {\em Proceedings of Machine Learning
  Research}, pages 1030--1038. {PMLR}, 2017.

\bibitem{Qi2012}
Q.~Qi, A.~Minturn, and Yang. Y.
\newblock {An Efficient Water-Filling Algorithm for Power Allocation in
  OFDM-Based Cognitive Radio Systems}.
\newblock {\em International Conference on Systems and Informatics}, 2012.

\bibitem{Welinder}
P.~{Welinder}, M.~{Welling}, and P.~{Perona}.
\newblock A lazy man's approach to benchmarking: Semisupervised classifier
  evaluation and recalibration.
\newblock In {\em 2013 IEEE Conference on Computer Vision and Pattern
  Recognition}, pages 3262--3269, 2013.

\bibitem{Druck}
G.~Druck and A.~McCallum.
\newblock {Toward Interactive Training and Evaluation}.
\newblock In {\em Proceedings of the 20th ACM International Conference on
  Information and Knowledge Management}, CIKM '11, page 947–956, New York,
  NY, USA, 2011. Association for Computing Machinery.

\bibitem{Yilmaz08Stratified}
E.~Yilmaz, E.~Kanoulas, and J.~A. Aslam.
\newblock A simple and efficient sampling method for estimating ap and ndcg.
\newblock In {\em Proceedings of the 31st Annual International ACM SIGIR
  Conference on Research and Development in Information Retrieval}, SIGIR '08,
  page 603–610, New York, NY, USA, 2008. Association for Computing Machinery.

\bibitem{pavlu07practical}
V.~Pavlu and J.~Aslam.
\newblock A practical sampling strategy for efficient retrieval evaluation.
\newblock {\em College of Computer and Information Science, Northeastern
  University}, 2007.

\bibitem{Aslam06Importance}
J.~A. Aslam, V.~Pavlu, and E.~Yilmaz.
\newblock A statistical method for system evaluation using incomplete
  judgments.
\newblock SIGIR '06, page 541–548, 2006.

\bibitem{MarchantRubinstein}
N.~G. Marchant and B.~I.~P. Rubinstein.
\newblock A general framework for label-efficient online evaluation with
  asymptotic guarantees.
\newblock {\em Arxiv}, 2006.06963, 2020.

\bibitem{pmlr-v80-nguyen18d}
P.~Nguyen, D.~Ramanan, and C.~Fowlkes.
\newblock {Active Testing: An Efficient and Robust Framework for Estimating
  Accuracy}.
\newblock In {\em ICML}, volume~80 of {\em Proceedings of Machine Learning
  Research}, pages 3759--3768. PMLR, 10--15 Jul 2018.

\bibitem{kossen2021active}
J.~Kossen, S.~Farquhar, Y.~Gal, and T.~Rainforth.
\newblock Active testing: Sample-efficient model evaluation.
\newblock {\em arXiv}, stat.ML(2103.05331), 2021.

\bibitem{lecun-mnisthandwrittendigit-2010}
Y.~LeCun and C.~Cortes.
\newblock {MNIST} handwritten digit database.
\newblock 2010.

\bibitem{Read_classifierchains}
J.~Read, B.~Pfahringer, G.~Holmes, and E.~Frank.
\newblock {Classifier Chains for Multi-label Classification}.
\newblock {\em Machine Learning Journal}, 85(3), 2011.

\end{thebibliography}

\appendix

\section{Importance Sampling\label{app:IS}}
We begin with a standard derivation of Importance Sampling being an unbiased estimator, here for the denominator term $\hat{x}$ in a estimator: 
\beq
\Fhat =\frac{\hat{x}}{\hat{y}}
\eeq
Both $\hat{x}$ and $\hat{y}$ are sums of independently distributed random variables. For large $M$ (and any value of $N$), $(\hat{x},\hat{y})$ will therefore be approximately jointly Gaussian distributed (Central Limit Theorem).  
Taking the expectation with respect to the Importance distribution, the Gaussian $p(\hat{x},\hat{y})$ has mean
\beq
\mu_x \equiv \ave{\hat{x}} = \frac{1}{MN}\sum_{i=1}^M \ave{\frac{\f{m_i}}{\q{m_i}}} =  \frac{1}{MN}\sum_{i=1}^M \sum_{n=1}^N \q{n} \frac{\avc{\f{n}}}{\q{n}}=\frac{1}{N}\sum_{n=1}^N \avc{\f{n}}
\eeq
Here $\avc{\cdot}$ is expectation with respect to the true label distribution $p(\ct_n|x_n)$. 
Similarly,
\beq
\mu_y \equiv \ave{\hat{y}} = \frac{1}{N}\sum_{n=1}^N \avc{\g{n}}
\eeq
Covariance elements of the Gaussian $p(\hat{x},\hat{y})$ can be computed using
\begin{align}
N^2M^2\ave{\hat{x}\hat{y}} &= \ave{\sum_{i,j} \frac{\f{m_i}\g{m_j}}{\q{m_i}\q{m_j}}}
= \ave{\sum_{i} \frac{\f{m_i}\g{m_i}}{\q{m_i}^2}} + \ave{\sum_{i\neq j} \frac{\f{m_i}\g{m_j}}{\q{m_i}\q{m_j}}}\\
&= M \sum_{n=1}^N \frac{\avc{\f{n}\g{n}}}{\q{n}} + (M^2-M) \sum_{n=1}^N \avc{\f{n}}\sum_{n=1}^N \avc{\g{n}}
\end{align}
so that the covariance between $\hat{x},\hat{y}$ is
\beq
\Sigma_{xy} = \frac{1}{M} \br{\frac{1}{N^2}\sum_{n=1}^N \frac{\avc{\f{n}\g{n}}}{\q{n}} - \mu_x\mu_y}
\eeq
Similarly,  
\beq
\Sigma_{xx} = \frac{1}{M} \br{\frac{1}{N^2}\sum_{n=1}^N \frac{\avc{\f{n}\f{n}}}{\q{n}} - \mu_x\mu_x}
\eeq
\beq
\Sigma_{yy} = \frac{1}{M} \br{\frac{1}{N^2}\sum_{n=1}^N \frac{\avc{\g{n}\g{n}}}{\q{n}} - \mu_y\mu_y}
\eeq
Since $\q{n}$ is typically $O(1/N)$, the mean elements are $O(1)$ whilst the covariance elements scale as $O(1/M)$ meaning that for large $M$ fluctuations from the mean will typically be small.  Writing $\hat{x}$ and $\hat{y}$ in terms of a mean-fluctuation decomposition, and expanding for a small fluctuation $\Delta$, 
\beq
\hat{\mathbb{F}} = \frac{\hat{x}}{\hat{y}} = \frac{\mu_x + \Delta_x}{\mu_y + \Delta_y}= \frac{1}{\mu_y}\frac{\mu_x + \Delta_x}{1 + \frac{\Delta_y}{\mu_y}}\approx \frac{1}{\mu_y}\br{\mu_x + \Delta_x}\br{1-\frac{\Delta_y}{\mu_y}}
\label{eq:delta:rep:app}
\eeq
Hence
\beq
\ave{\hat{\mathbb{F}}} = \frac{\mu_x}{\mu_y}\br{1 - \frac{\Sigma_{xy}}{\mu_x\mu_y}} + O(1/M^2)
\label{eq:av:f:app}
\eeq
The expected metric therefore tends to
\beq
\FMinf \equiv \frac{\mu_x}{\mu_y} = \frac{\sum_{n=1}^N \avc{f_n}}{\sum_{n=1}^N \avc{g_n}}
\label{eq:FMinf:app}
\eeq
as the number of IS samples $M\rightarrow\infty$. This is the exact expected value of the metric calculated on the test set and holds for any $N$. We wish to find an estimator that accurately matches this ideal value $\Fp$.  From \eqref{eq:av:f:app} we see that $\Fhat$ is a consistent, but biased (with bias $O(1/M)$) estimator of $\Fp$.

As $N\rightarrow\infty$, the means $\mu_x$, $\mu_y$ tend to (from the law of large numbers)
\beq
\mu_x \rightarrow \ave{f(\cp,\ct)}, \hcm \mu_y \rightarrow \ave{g(\cp,\ct)}, 
\eeq
where the expectation is with respect to the joint $p(\cp,\ct,x)=p_d(\cp|x)p(\ct|x)p(x)$ for decision distribution\footnote{Here $p_d(\cp|x)$ represents the decision probability as a function of the model $p(\cp|x)$. For example, if we use the model $p(\cp|x)$ to form a deterministic classifier by thesholding, then $p_d(\cp=1|x)=\ind{p(\cp|x)>\theta}$ for some user defined threshold $\theta\in\sq{0,1}$.} $p_d(\cp|x)$. Hence, as both $M$ and $N$ become large, the estimator $\Fhat$ converges to the true test metric (evaluated on infintely many test points $N$)
\beq
\Fhat \rightarrow \frac{\ave{f(\cp,\ct)}}{\ave{g(\cp,\ct)}}\equiv \FNinf
\eeq
$\Fhat$ is therefore also a consistent (but biased) estimator of $\FNinf$.  We note that estimating the infinite testset size ($N{}\rightarrow{}\infty$) performance $\FNinf$ is not our aim. Rather, our aim is to estimate the performance $\FMinf$ on the given, finite $N$ dataset. In our scenario, if we could evaluate the test performance on all $N$ datapoints, we would be able to compute the exact metric (for a deterministic classifier $p(\ct_n|x_n)$).

\subsection{Error Calculation \label{sec:opt:is:error}}
The squared error between the finite $M$ estimator $\Fhat$ and infinite $M$ limit $\FMinf$  measures the error from IS in approximating the finite $N$ metric $\Fp$. From \eqref{eq:delta:rep:app},  
\begin{align}
\ave{\br{\Fhat  - \FMinf}^2} &=\frac{1}{\mu_y^2}\ave{\br{\Delta_x - \FMinf\Delta_y}^2}
=\frac{1}{\mu_y^2}\br{\Sigma_{xx} - 2\FMinf\Sigma_{xy} + \FMinf^2\Sigma_{yy}}
\end{align}
where $\Fp = \mu_x/\mu_y$. 
%
We can write $M\br{\Sigma_{xx} - 2\FMinf\Sigma_{xy} + \FMinf^2\Sigma_{yy}}$ as
\begin{align}
&= \sum_n\frac{1}{N^2} \frac{\avc{\f{n}^2}}{\q{n}} - \mu_x^2 -2\FMinf\br{\frac{1}{N^2}\sum_n \frac{\avc{\f{n}\g{n}}}{\q{n}}-\mu_x\mu_y}+\FMinf^2\br{\frac{1}{N^2}\sum_n \frac{\avc{\g{n}^2}}{\q{n}} - \mu_y^2}\\
&= \frac{1}{N^2}\sum_n \frac{\avc{\br{\f{n} - \FMinf \g{n}}}^2}{\q{n}} - \br{\mu_x - \FMinf\mu_y}^2 \\
&= \frac{1}{N^2}\br{\sum_n \frac{\avc{\br{\f{n} - \FMinf \g{n}}^2}}{\q{n}}}. 
\end{align}

The squared error between the finite $M$ estimator $\Fhat$ and infinite $M$ limit $\FMinf$
is
\beq
\ave{\br{\Fhat  - \FMinf}^2} =\frac{1}{\mu_y^2}\ave{\br{\Delta_x - \FMinf\Delta_y}^2} + O(1/M^2)
\eeq
Hence, to leading order in $1/M$
\beq
\ave{\br{\Fhat  - \FMinf}^2} 
=\frac{1}{M\mu_y^2}\br{\frac{1}{N^2}\sum_{n=1}^N \frac{\avc{\br{f_n - \FMinf g_n}^2}}{q_n} - \br{\mu_x-\FMinf\mu_y}^2}
\eeq
Since $\Fp=\mu_x/\mu_y$, we note that the term $\mu_x-\FMinf\mu_y$ is zero.
Taking the expectation $\avc{\cdot}$ with respect to the true generating probability $p(\ct_n|x_n)$ we obtain
\beq
\ave{\br{\Fhat  - \FMinf}^2}= \frac{1}{MN^2\mu_y^2}\sum_{n=1}^N \frac{h_n^2}{q_n}
\label{eq:is:error:app}
\eeq
where 
\begin{align}
h_n^2 &=  \avc{\br{f_n - \FMinf g_n}^2}\label{eq:h:gen:def}\\
& = p(c_n^t=1|x_n) \br{f(\cp_n,\ct_n=1) -\FMinf g(\cp_n,\ct_n=1)}^2\nonumber\\
&\hspace{3.5cm}+p(\ct_n=0|x_n) \br{f(\cp_n,\ct_n=0) -\FMinf g(\cp_n,\ct_n=0)}^2     
\end{align}

\subsection{Optimal Sampling Distribution\label{sec:opt:is}}

The sampling distribution that minimises the variance can be calculated using the Lagrangian
\beq
\sum_{n=1}^N \frac{\h{n}^2}{\q{n}} - \lambda\br{\sum_{n=1}^N \q{n}-1}.
\eeq
Differentiating with respect to $\q{n}$ and equating to zero gives the optimal choice as
\beq
\q{n} = \frac{|\h{n}|}{\sum_{n=1}^N |\h{n}|}.
\label{eq:opt:is:dist}
\eeq
Clearly, it isn't possible to construct this optimal estimator in practice, since this would require us to know the true metric $F$. Hence, in practice, we use
\beq
\q{n} = \frac{\tilde{h}_n}{\sum_{n=1}^N \tilde{h}_n}
\eeq
for some approximation $\tilde{h}_n$ of $\h{n}$.

\section{Bernoulli Sampling\label{app:bs}}

We again consider the problem of approximating
\beq
\frac{\sum_{n=1}^N f_n}{\sum_{n=1}^N g_n}
\eeq
The Bernoulli Sampler \cite{CSS} which can be used to form an estimator
\beq
\Fhat = \frac{\hat{x}}{\hat{y}}
\label{eq:fhat:bs:app}
\eeq
where
\beq
\hat{x} = \frac{1}{N}\sum_{n=1}^N \frac{s_n}{b_n}f_n , \ocm \hat{y} = \frac{1}{N}\sum_{n=1}^N \frac{s_n}{b_n}g_n 
\label{eq:xyhat:bs:app}
\eeq
Since $\hat{x}$ and $\hat{y}$ are sums of independently generated random variables, for large $N$, $p(\hat{x},\hat{y})$ will be approximately Gaussian distributed with mean 
\beq
\mu_x = \frac{1}{N}\sum_n \frac{\ave{s_n f_n}}{b_n}=\frac{1}{N}\sum_n \avc{f_n}, \ocm \mu_y = \frac{1}{N}\sum_n \avc{g_n},
\eeq
where $\ave{s_n f_n}=\ave{s_n}\av{f_n}=b_n\av{f_n}$. Here we used the fact that for a 0/1 binary variable $\ave{s_n}=p(s_n=1|b_n)=b_n$; as for IS, $\av{\cdot}$ denotes expectation with to a stochastic true classifier. The covariance elements are also straightforward to calculate:
\begin{align}
\Sigma_{xy} &= \frac{1}{N^2}\sum_{n=1}^N \br{\frac{\avc{f_ng_n}}{b_n}-\avc{f_n}\avc{g_n}},\\
\Sigma_{yy} &= \frac{1}{N^2}\sum_{n=1}^N \br{\frac{\avc{g_ng_n}}{b_n}-\avc{g_n}\avc{g_n}},\\
\Sigma_{xx} &= \frac{1}{N^2}\sum_{n=1}^N \br{\frac{\avc{f_nf_n}}{b_n}-\avc{f_n}\avc{f_n}}
\end{align}
Since the covariance elements are $O(1/N)$ compared to the $O(1)$ mean elements,  fluctuations from the mean are typically small and we can use the  mean-fluctuation decomposition:
\beq
\hat{\mathbb{F}} = \frac{\hat{x}}{\hat{y}} = \frac{\mu_x + \Delta_x}{\mu_y + \Delta_y}= \frac{1}{\mu_y}\frac{\mu_x + \Delta_x}{1 + \frac{\Delta_y}{\mu_y}}\approx \frac{1}{\mu_y}\br{\mu_x + \Delta_x}\br{1-\frac{\Delta_y}{\mu_y}}
\label{eq:delta:rep:two:app}
\eeq
Hence
\beq
\ave{\hat{\mathbb{F}}} = \frac{\mu_x}{\mu_y}\br{1 - \frac{\Sigma_{xy}}{\mu_x\mu_y}} + O(1/N^2)= \frac{\mu_x}{\mu_y} + O(1/N)
\eeq
In the deterministic true classifier setting, the bias of this estimator is therefore approximately
\beq
\frac{1}{N^2\mu_y^2}\sum_{n=1}^N{f_ng_n}\br{\frac{1}{b_n}-1}
\label{eq:bias:det:app}
\eeq
As $N\rightarrow\infty$, the means $\mu_x$, $\mu_y$ tend to (from the law of large numbers)
\beq
\mu_x \rightarrow \ave{f(\cp,\ct)}, \hcm \mu_y \rightarrow \ave{g(\cp,\ct)}, 
\eeq
where the expectation is with respect to the joint $p(\cp,\ct,x)=p_d(\cp|x)p(\ct|x)p(x)$.  Hence, as $N$ becomes large, the estimator $\Fhat$ converges to the value
\beq
\Fhat \rightarrow \frac{\ave{f(\cp,\ct)}}{\ave{g(\cp,\ct)}}\equiv \FNinf
\eeq
As for IS, the BS estimator $\Fhat$ is therefore a consistent (but biased) estimator of $\FNinf$. 

The expected squared error of the estimator in approximating the metric on the given dataset is then (to leading order in $1/N$)
\begin{align}
\ave{\br{\Fhat  - \Fp}^2} &\approx\frac{1}{\mu_y^2}\ave{\br{\Delta_x - \mathbb{F}\Delta_y}^2}=\frac{1}{\mu_y^2}\br{\Sigma_{xx} - 2\Fp\Sigma_{xy} + \Fp^2\Sigma_{yy}}\\
&= \frac{1}{N^2\mu_y^2}\sum_{n=1}^N\br{\frac{1}{b_n}\avc{\br{f_n - \Fp g_n}^2} - \br{\avc{f_n}-\Fp\avc{g_n}}^2}
\label{eq:var:est:two:app}
\end{align}
In the above $\Fp$ is the true value of the metric on the given finite $N$ testset, defined in \eqref{eq:FMinf:app}. In the deterministic true classifier setting, this is
\beq
\ave{\br{\Fhat  - \Fp}^2} \approx \frac{1}{N^2\mu_y^2}\sum_{n=1}^N{\br{f_n - \Fp g_n}^2}\br{\frac{1}{b_n}-1}
\label{eq:var:est:det:app}
\eeq


\subsection{Optimal sampling probabilities\label{sec:BS:optb}}

The objective is to minimise the function
\beq
\sum_{n=1}^N \frac{\h{n}^2}{b_n}
\eeq
subject to the constraints $\sum_n b_n=M$ and $0< b_n \leq 1$. We note first that the constraints are convex and that the function is convex in the feasible set. Therefore the objective function has a unique minimum value. 

Minimising the variance while keeping the expected number of samples fixed to $M$ requires solving the Lagrangian
\beq
\sum_{n=1}^N \frac{\h{n}^2}{b_n} + \lambda \br{M-\sum_n b_n}.
\eeq
Since we have the requirement $0\leq b_n \leq 1$ we parameterise
\beq
b_n= e^{-\gamma_n^2}.
\eeq
Taking the derivative of the Lagrangian wrt $\gamma_n$ gives
\beq
\gamma_n\br{\h{n}^2/b_n-\lambda b_n} =0,
\eeq
so that either $\gamma_n=0$ ($b_n=1$) or,
\beq
\frac{\h{n}^2}{b_n} = \lambda b_n \Rightarrow b_n = \frac{\h{n}}{\sqrt{\lambda}}
\label{eq:b:opt:prop}
\eeq
Defining binary indicators $i_n\in\cb{0,1}$ we can write these two possibilities as 
\beq
b_n = i_n + \kappa(i)(1-i_n)\h{n}
\label{eq:b:def}
\eeq
for some the function $\kappa(i)$.  Since $\sum_n b_n=M$ we have
\beq
M = \sum_n i_n + \kappa(i) \sum_n \br{1-i_n}\h{n}
\eeq
so that
\beq
\kappa(i) \equiv \frac{M - \sum_n i_n}{\sum_n \br{1-i_n}\h{n}}
\label{eq:kappa}
\eeq
Plugging this back into the objective, we have
\beq
E(i) \equiv \sum_n \frac{\h{n}^2}{b_n}  = \sum_n \h{n}^2 \br{i_n + (1-i_n)\frac{1}{\kappa(i) \h{n}}}=\sum_n i_n h_n^2 + \frac{1}{M-\sum_n i_n}\br{\sum_n(1-i_n)h_n}^2
\label{eq:error:inds}.
\eeq

Since the highest contributions to the error arise from the largest values of $\h{n}$, the values of $b_n$ must be ordered according to decreasing values of $\h{n}$. That is, the largest values of $b_n$ are associated with the largest values of $\h{n}$. We note also that the objective is convex on the feasible set $0\leq b_n \leq 1$. This means that the first valid solution we find is guaranteed to equal the global minimal value. A simple $O(N)$ algorithm to find the global minimum is given in \algref{alg:b:op}. 


\begin{algorithm}[t]
	\caption{Optimal Bernoulli Distribution. Find $b_n\in\sq{0,1}$ that minimise $\sum_{n=1}^N h_n^2/b_n$ subject to $\sum_{n=1}^N b_n=M$\label{alg:b:op}} 
	\begin{algorithmic}[1]
	\State $H=\sum_{n=1}^N h_n$
	\State Order the absolute deviations $\h{n}$ from highest to lowest:
$\h{\omega_1}\geq \h{\omega_2},\ldots\geq \h{\omega_N}$
	\If {$M*h_{\omega_1}\leq H$} 
	\State For each $n\in\cb{1,\ldots,N}$, set $b_n= Mh_n/H$ and return $b$ as the solution
	\Else
	\State $S=0$
    \For {$j = 1 : N$}
    \State $S = S + h_{\omega_j}$
	\State $G = h_{\omega_j}*(M-j)/(H-S)$
	\If {$G\leq 1$}
	\State $b_{\omega_1:\omega_{j-1}}=1$
	\For {$r = j : N$} 
	\State $b_{\omega_{r}}=h_{\omega_r}*(M-j)/(H-S)$
	\EndFor
	\State Return $b$ as the solution
	\EndIf
	\EndFor
	\EndIf
	\end{algorithmic} 
\end{algorithm}


\section{Importance Sampling versus Bernoulli Sampling\label{sec:is:vs:bs}}

From \eqref{eq:is:error:app} the expected squared error for IS is, using the optimal IS distribution \eqref{eq:opt:is:dist}
\beq
E_{IS}\equiv \ave{\br{\Fhat  - \FMinf}^2}_{IS}= \frac{1}{M_{IS}N^2\mu_y^2}\sum_{n=1}^N \frac{h_n^2}{q_n}
=\frac{1}{M_{IS}\mu_y^2}\br{\frac{1}{N}\sum_{n=1}^N h_n}^2
\label{eq:is:error:two}
\eeq
For BS we do not have a simple expression for the optimal setting of the sampling parameters $b_n$. However, when $M_{BS}$ is small compared to $N$, the $b_n$ typically do not saturate to 1 and that therefore, from \eqref{eq:b:opt:prop}, $b_n\propto h_n$. Using the constraint $\sum_{n=1}^N b_n=M_{BS}$, provided that all $b_n<1$, we have $b_n=M_{BS} h_n/\sum_n h_n$. The expected squared error, from \eqref{eq:var:est:two} in the deterministic true-classifier case is then
\beq
E_{BS}\equiv \ave{\br{\Fhat  - \Fp}^2}_{BS} = \frac{1}{N^2\mu_y^2}\sum_{n=1}^N \br{\frac{1}{b_n}-1}h_n^2= \frac{1}{M_{BS}\mu_y^2}\br{\frac{1}{N}\sum_{n=1}^N h_n}^2 - \frac{1}{N^2\mu_y^2}\sum_{n=1}^N h_n^2
\label{eq:var:est:three}
\eeq
meaning that the expected BS error is less than the expected IS error if the same number of samples $M_{IS}=M_{BS}$ is used. 

However, this question is made more complex by the fact that, in a deterministic true classifier setting, repeated use of the same sample does not add to the labelling cost. In IS, due to resampling, the inclusion probability in $M$ samples is
\beq
\pi_n = 1 - (1-q_n)^M
\eeq
so that the expected number of unique samples in $M$ drawn samples is, 
\beq
\sum_n \pi_n = N - \sum_n  (1-q_n)^M
\eeq
We therefore suggest the sample equivalence 
\beq
M_{BS} =  N- \sum_n  (1-q_n)^{M_{IS}}
\label{eq:is:bs:equiv}
\eeq
Using the simple bound
\beq
\frac{1}{N}\sum_n h_n^2 \geq \br{\frac{1}{N}\sum_n h_n}^2
\eeq
we can write
\beq
E_{BS}\leq \br{\frac{1}{M_{BS}}-\frac{1}{N}}M_{IS}E_{IS} 
\eeq
For small $M_{IS}$ and large $N$ (so that each $q_n$ is small), from \eqref{eq:is:bs:equiv}
\beq
M_{BS} \approx N - \sum_n ( 1 - M_{IS}q_n) = N - N + M_{IS} = M_{IS}
\eeq
so that (for small $M_{IS}/N$)
\beq
E_{BS}\leq \br{1-\frac{M_{IS}}{N}}E_{IS} 
\eeq
In the limit of small $M_{IS}$,  $E_{BS}\approx E_{IS}$ (as borne out by our experiments). However, as $M_{IS}$ increases BS starts to significantly outperform IS.  This shows that there is no reason to prefer IS over BS since BS will have at least as good performance as IS (even though they have equivalent performance in the very low sample limit).

The difference between IS and BS will become more significant when the BS weights start to saturate to 1. This happens when  (from \algref{alg:b:op}) 
\beq
h^* \geq \frac{N}{M_{BS}}\bar{h}
\eeq
where $h^*$ is the maximum of the deviations $h_1,\ldots, h_N$ and $\bar{h}$ is the average of the deviations.

\section{Multi-Labels\label{sec:multi}}

In the multi-labelling scenario, an input $x_n$ can have multiple labels $c_n\in\cb{1,\ldots,C}$; for example, a sentence might be classed as "upbeat", "humorous" and "about cats". A standard way to address this is to consider a set of binary classifiers, each predicting the presence/absence of each of the $C$ attributes -- sometimes called the binary relevance approach\cite{Read_classifierchains}. For simplicity, we assume the $C$ binary classifiers are independent, conditioned on the input $x_n$. 

Writing $\cp_n = \br{v_1,\ldots,v_C}$ where each $v_k\in\cb{0,1}$ the model prediction is then given by
\beq
p(\cp_n|x_n) = \prod_{k=1}^C p(v_k=1|x_n)
\eeq
where  $p(v_k=1|x_n)$ is a user provided binary classifier for each class.

\subsection{Micro $F_1$}

For this setting, a common metric is the Micro-$F_1$ score defined as 
\beq
\FalphaMicro \equiv \frac{\sum_{n=1}^N\sum_{i=1}^C \ind{\cp_n=i,\ct_n=i}}{\sum_{n=1}^N\sum_{i=1}^C \br{\alpha\ind{\cp_n=i} + (1-\alpha)\ind{\ct_n=i}}}
\eeq
where $\cp_n=i$ means $\vp_{n,i}=1$ and $\cp_n\neq i$ means $\vp_{n,i}=0$. 

This is of the general form $\sum_n f_n/\sum_n g_n$ where
\beq
f_n = \sum_{i=1}^C \ind{\cp_n=i,\ct_n=i}, \hcm g_n  = \sum_{i=1}^C \br{\alpha\ind{\cp_n=i} + (1-\alpha)\ind{\ct_n=i}}
\eeq
We may then use the same strategy as in \eqref{eq:pred:approx} to form an approximation to the true classifier using
\beq
p_a(\ct_n|x_n) = \prod_{k=1}^C p_a(v_k=1|x_n)
\label{eq:multi:pa}
\eeq
where
\beq
p_a(v_k=1|x_n) = \lambda p(v_k=1|x_n) + (1-\lambda)0.5
\eeq

The previous Optimal IS and Optimal BS theory still holds in this setting and to calculate the optimal sampling distributions we need to calculate the expected squared deviation 
\beq
h_n^2 =  \avc{\br{f_n - \Fp g_n}^2} 
\eeq
where the quantities are approximated by taking expectation with respect to $p_a(\ct_n|x_n)$.   This is a straightforward calculation, see \appref{app:multi}. This can then be used to define the Optimal Importance and Bernoulli Samplers, as before.  We estimate the error in the resulting estimator of $\FalphaMicro$ by using the same approach as for IS and BS.


\subsection{Micro $F_1$ Deviation\label{app:multi}}

Writing $h_n(v) = \sum_{k=1}^C h(v_{nk})$, where $h(v_{nk}) = \f{}(\vp_{nk},\vt_{nk}) - \Fp  \g{}(\vp_{nk},\vt_{nk})$
a straightforward calculation gives the expectation $h_n^2\equiv \av{h_n(v)^2}$
as
\begin{align}
h_n^2 
 &=  \sum_k p_a(v_k=0|x_n)\h{}^2(\vp_{nk},\vt_k=0) + p_a(v_k=1|x_n)\h{}^2(\vp_{nk},\vt_k=1)\\
 &-  \sum_k \br{p_a(v_k=0|x_n)\h{}(\vp_{nk},\vt_k=0) + p_a(v_k=1|x_n)\h{}(\vp_{nk},\vt_k=1)}^2\\
 &+  \br{\sum_k p_a(v_k=0|x_n)\h{}(\vp_{nk},\vt_k=0) + p_a(v_k=1|x_n)\h{}(\vp_{nk},\vt_k=1)}^2
\label{eq:h:def:micro}
\end{align}

\subsection{Macro $F_1$}

We first compute the true positive, false positive and false negative rates for each class:
\begin{align}
TP_i &= \frac{1}{N}\sum_{n=1}^N \ind{\cp_n=i,\ct_n=i}\\
FP_i &= \frac{1}{N}\sum_{n=1}^N \ind{\cp_n=i,\ct_n\neq i}\\
FN_i &= \frac{1}{N}\sum_{n=1}^N \ind{\cp_n\neq i,\ct_n=i}
\end{align}
Then the macro Precision and Recall are defined as
\beq
P =  \frac{1}{C}\sum_{i=1}^C \frac{TP_i}{TP_i+FP_i}, \hcm R =  \frac{1}{C}\sum_{i=1}^C \frac{TP_i}{TP_i+FN_i}, 
\eeq
and we finally define the the Macro $F_1$ score as
\beq
\F = \frac{2PR}{P+R}
\eeq
This cannot be written in the form $\sum_n f_n/\sum_n g_n$ and our previous theory cannot be directly applied. For notational simplicity, we define
\beq
\bar{a}_i \equiv TP_i, \hcm \bar{b}_i \equiv FP_i, \hcm \bar{c}_i\equiv FN_i
\eeq
and define a BS sampling approximation for each quantity as
\begin{align}
a_i &= \frac{1}{N}\sum_{n=1}^N \frac{s_n}{b_n}\ind{\cp_n=i,\ct_n=i}\\
b_i &= \frac{1}{N}\sum_{n=1}^N \frac{s_n}{b_n}\ind{\cp_n=i,\ct_n\neq i}\\
c_i &= \frac{1}{N}\sum_{n=1}^N \frac{s_n}{b_n}\ind{\cp_n\neq i,\ct_n=i}
\end{align}
We can make use of the fact that these estimators are sums of independent random variables and as such as jointly Gaussian distributed.  
Then the macro $F_1$ is a function of these variables:
\beq
\frac{C}{2}\F(a,b,c) = \frac{\sum_i a_i/(a_i+b_i) \sum_i a_i/(a_i+c_i)}{\sum_i a_i/(a_i+b_i) + \sum_i a_i/(a_i+c_i)}  
\eeq
By the Central Limit Theorem, $a,b,c$ will concentrate around their average values for large $N$ and we can write (see \appref{app:f1:macro})
\beq
\F(a,b,c) \approx \frac{1}{N^2}\sum_{n=1}^N \frac{h_n^2}{b_n}
\eeq
where
\begin{multline}
\h{n}^2\equiv \sum_{i=1}^C \ind{\cp_n=i}\av{\ind{\ct_n=i}}\br{\partial_{a_i}\F}^2 +  \ind{\cp_n=i}\av{\ind{\ct_n\neq i}}\br{\partial_{b_i}\F}^2\\
+ \ind{\cp_n\neq i}\av{\ind{\ct_n=i}}\br{\partial_{c_i}\F}^2
\end{multline}
We evaluate the derivative of $\F$ at $a=\avp{p_a}{a}$, $b=\avp{p_a}{b}$, $c=\avp{p_a}{c}$  where the expectation is with respect to $p_a$, \eqref{eq:multi:pa}. This then enables us to compute an approximation to $\h{n}^2$ which we use to define the Optimal BS, using \algref{alg:bs} as usual.

\begin{figure}[t]
\centering
\includegraphics[width=0.45\textwidth]{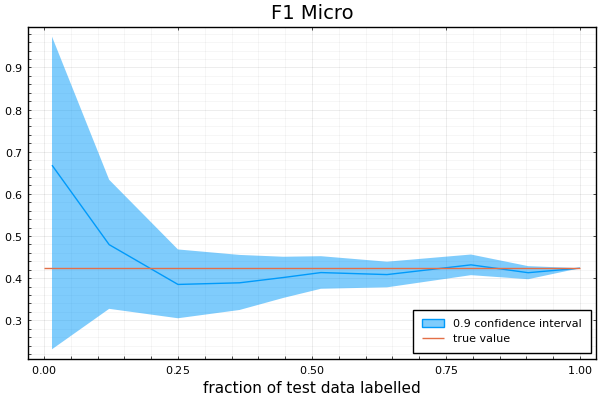}
\includegraphics[width=0.45\textwidth]{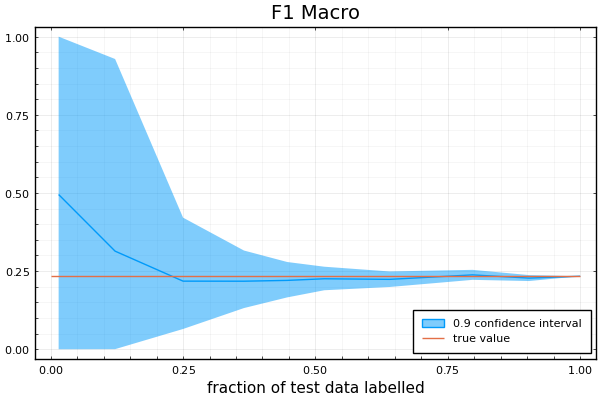}\\
\includegraphics[width=0.45\textwidth]{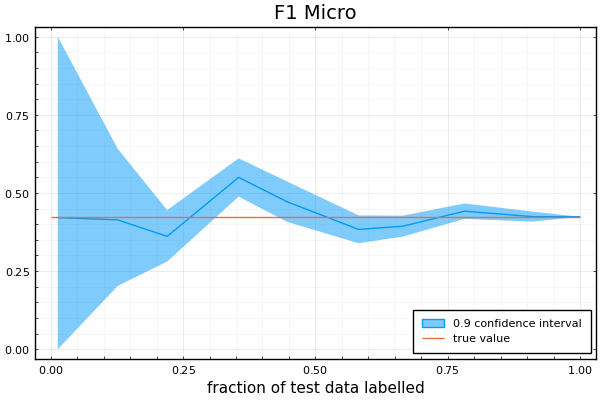}
\includegraphics[width=0.45\textwidth]{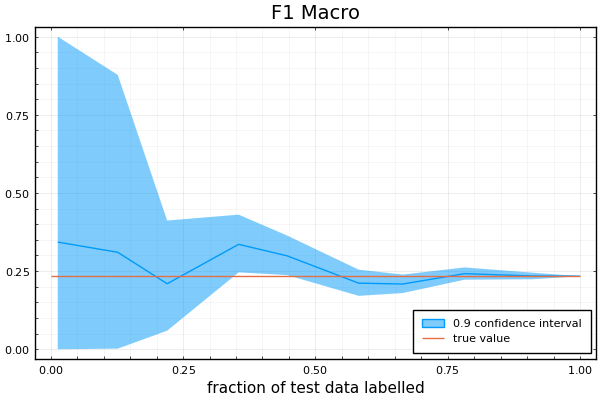}
\caption{Using Bernoulli Sampling to estimate the micro and macro $F_1$  score for the Twitter Toxic Comments classification problem. First row: Bernoulli samples were drawn to optimise the micro $F_1$ score. Second row: Bernoulli samples were drawn to optimise the macro $F_1$ score. \label{fig:multi:bs}}
\end{figure}

\subsection{Macro $F_1$ Deviation\label{app:f1:macro}}

Then the macro $F_1$ is given by
\beq
\frac{C}{2}\F(a,b,c) = \frac{\sum_i a_i/(a_i+b_i) \sum_i a_i/(a_i+c_i)}{\sum_i a_i/(a_i+b_i) + \sum_i a_i/(a_i+c_i)}  
\eeq
By the Central Limit Theorem, $a,b,c$ will concentrate around their average values for large $N$ and we can write 
\beq
\F(a,b,c) \approx \F(\bar{a},\bar{b},\bar{c}) + \Delta_a\trans \partial_a \F+ \Delta_b\trans \partial_b \F+ \Delta_c\trans \partial_c \F 
\eeq
Using summation convention on repeated indices $i$ and the fact that classes are conditionally independent, to leading order:
\begin{multline}
\ave{\br{\F(a,b,c) - \F(\bar{a},\bar{b},\bar{c}}^2} = \br{\partial_{a_i}\F}^2\ave{\Delta a_i^2} + 
\br{\partial_{b_i}\F}^2\ave{\Delta b_i^2} + \br{\partial_{c_i}\F}^2\ave{\Delta c_i^2}\\
+ 2 \br{\partial_{a_i}\F\partial_{b_i}\F}\ave{\Delta a_i\Delta b_i}
+ 2 \br{\partial_{a_i}\F\partial_{c_i}\F}\ave{\Delta a_i\Delta c_i}
+ 2 \br{\partial_{b_i}\F\partial_{c_i}\F}\ave{\Delta b_i\Delta c_i}
\end{multline}
\beq
\ave{\Delta a_i^2}=\frac{1}{N^2}\sum_{n=1}^N \ind{\cp_n=i}\av{\ind{\ct_n=i}}\br{\frac{1}{b_n}-1}
\eeq
\beq
\ave{\Delta b_i^2}=\frac{1}{N^2}\sum_{n=1}^N \ind{\cp_n=i}\av{\ind{\ct_n\neq i}}\br{\frac{1}{b_n}-1}
\eeq
\beq
\ave{\Delta c_i^2}=\frac{1}{N^2}\sum_{n=1}^N \ind{\cp_n\neq i}\av{\ind{\ct_n=i}}\br{\frac{1}{b_n}-1}
\eeq
\beq
\ave{\Delta a_i\Delta b_i}=-\frac{1}{N^2}\sum_{n=1}^N \ind{\cp_n=i}\av{\ind{\ct_n=i}}\av{\ind{\ct_n\neq i}}
\eeq
\beq
\ave{\Delta a_i\Delta c_i}=\ave{\Delta b_ic_i}=0
\eeq
We can therefore write the Bernoulli probability $b$-dependence of the expected squared error as
\beq
\sum_{n=1}^N\frac{\h{n}^2}{b_n}
\eeq
where
\begin{multline}
\h{n}^2\equiv \ind{\cp_n=i}\av{\ind{\ct_n=i}}\br{\partial_{a_i}\F}^2
+  \ind{\cp_n=i}\av{\ind{\ct_n\neq i}}\br{\partial_{b_i}\F}^2\\
+\ind{\cp_n\neq i}\av{\ind{\ct_n=i}}\br{\partial_{c_i}\F}^2
\end{multline}

\section{Online Bernoulli Sampling\label{app:online}}

Unlike Importance Sampling, in which a data index $n$ can be selected one at a time, by construction in Bernoulli Sampling a \emph{collection} of indices $\sett{N}=\cb{n : s_n = 1}$ are sampled. However, it is still possible to form an online Bernoulli process in which at each sampling round, a subset of indices are drawn, based on previously drawn indices. For example, consider a simple metric of the form $\sum_{n=1}^N f_n$. In the first round of Bernoulli Sampling, we have weights $b^1=b^1_1,\ldots,b^1_N$ and samples $s^1_1,\ldots,s^1_N$ and thus indices $\sett{N}_1$ that correspond to $s^1_n=1$ from this Bernoulli distribution.  We would then like to use these sampled values to inform the second stage of Bernoulli sampling. We write the first round estimator as 
\beq
\Fhat_1 = \sum_{n=1}^N \frac{s^1_{n}}{b^1_n} f_{n} = \sum_{n\in \sett{N}_1} \frac{f_n}{b^1_n}
\eeq
In the second round, we do not wish to redraw the set $\sett{N}_1$ and the weights $b^2$ will therefore depend on the previously drawn indices $\sett{N}_1$. We can then form a new estimator
\beq
\Fhat_2 = \sum_{n=1}^N \frac{s^2_{n}}{b^2_n} f_{n}
\eeq
Since $\Fhat_2$ does not contain the previously sampled indices, it cannot be an unbiased estimator of $\F$. A simple way to compensate for this (analogous to the PURE  estimator \cite{kossen2021active} for IS) is to write
\beq
\Fhat_{1:2} = \frac{1}{2}\Fhat_1 + \frac{1}{2}\br{\sum_{n\in\sett{N}_1} f_n + \Fhat_2} 
\eeq
Taking the expectation with respect to the two rounds $p(s^2|s^1)p(s^1)$ we have
\begin{align}
\avp{s}{\Fhat_{1:2}} &= \frac{1}{2}\avp{s_1}{\Fhat_1} + \frac{1}{2}\avp{s_1}{{\sum_{n\in\sett{N}_1} f_n + \avp{s_2|s_1}{\Fhat_2}}}\\ 
&= \frac{1}{2}\avp{s_1}{\Fhat_1} + \frac{1}{2}\avp{s_1}{{\sum_{n\in\sett{N}_1} f_n + \sum_{n\in\sett{N}\setminus\sett{N}_1} f_n}}\\
&= \frac{1}{2}\F + \frac{1}{2}\avp{s_1}\F\\
&= \F
\end{align}
More generally, for the $r^{th}$ round of Bernoulli Sampling (designed so that it cannot select any of the previously selected datapoints) we write an estimator for the sum over the remaining non-sampled indices as
\beq
\Fhat_r = \sum_{n=1}^N \frac{s_n}{b_n^r}f_n
\eeq 
Then writing $\sett{N}_{1:R-1}$ for all indices that have been sampled in the previous $R-1$ rounds, we can form an unbiased estimator of $\F$ using
\beq
\Fhat_{1:R} = \frac{1}{R}\sum_{r=1}^R\br{\sum_{n\in\sett{N}_{1:r-1}}f_n  + \Fhat_r}, \ocm \sett{N}_0=\emptyset
\eeq
We note that the choice of distribution $p(s^r|s^{1:r-1})$ can depend on an updated online estimate of the metric, and the estimator $\Fhat_{1:R}$ remains unbiased.  However, the theoretical and empirical properties of such an online Bernoulli Sampler and its implementation in the estimation of general metrics $\sum_n f_n/\sum_n g_n$ is left for future study.


\section{Additional Results\label{sec:add:results}}

We show here additional results for the experiments presented in the main text. 

\begin{figure}
\centering
\includegraphics[width=0.31\textwidth]{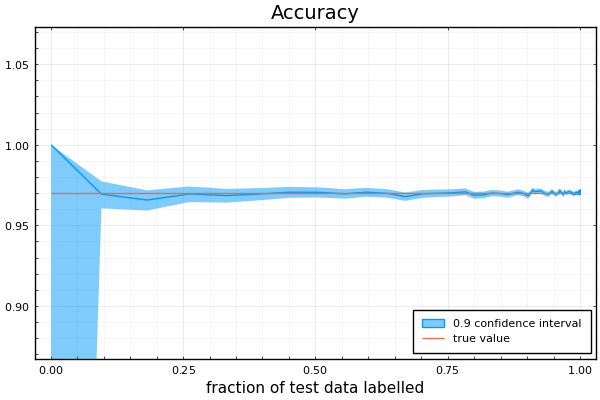}\includegraphics[width=0.3\textwidth]{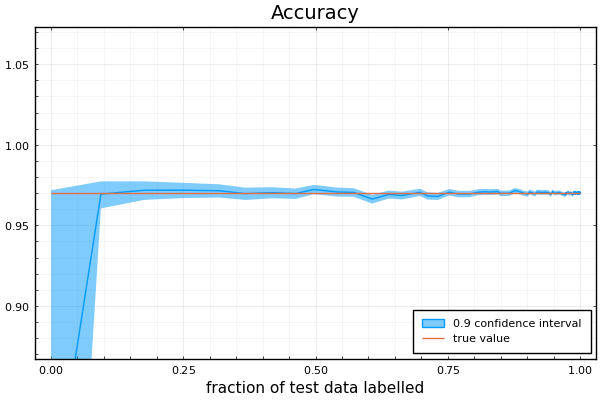}\includegraphics[width=0.3\textwidth]{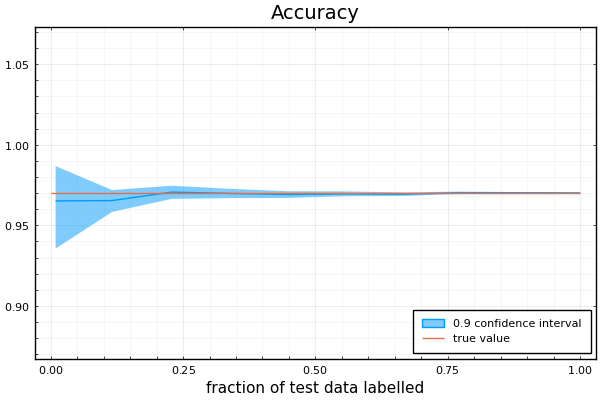}
\includegraphics[width=0.31\textwidth]{US_F1}\includegraphics[width=0.3\textwidth]{IS_F1}\includegraphics[width=0.3\textwidth]{BS_F1}
\includegraphics[width=0.31\textwidth]{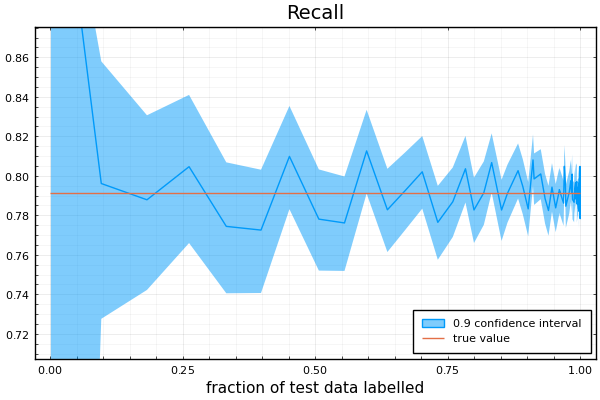}\includegraphics[width=0.3\textwidth]{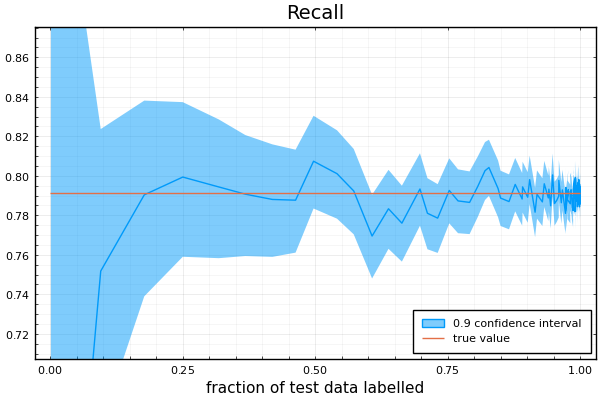}\includegraphics[width=0.31\textwidth]{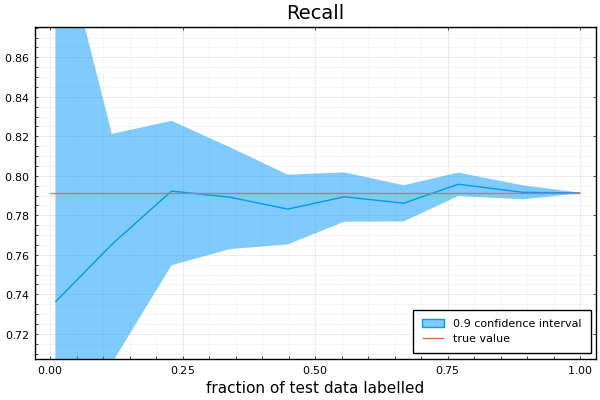}
\includegraphics[width=0.31\textwidth]{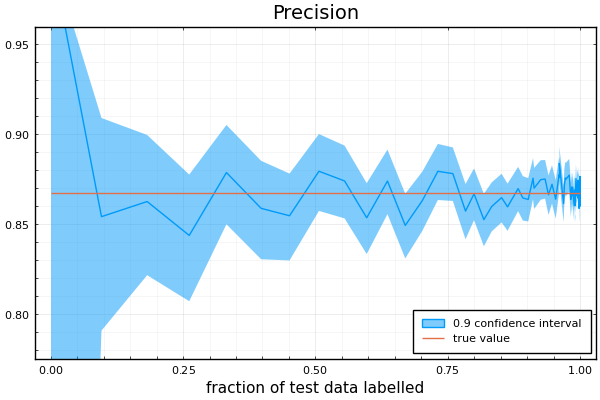}\includegraphics[width=0.3\textwidth]{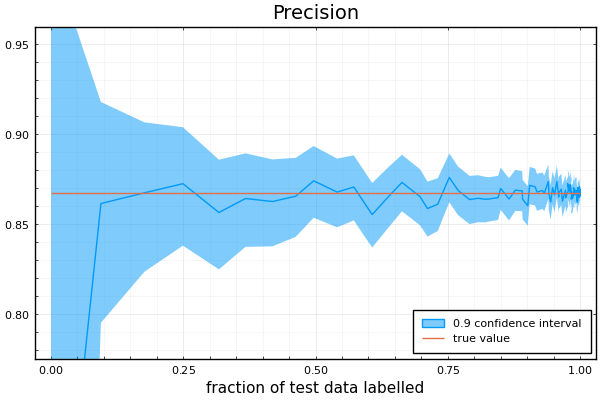}\includegraphics[width=0.31\textwidth]{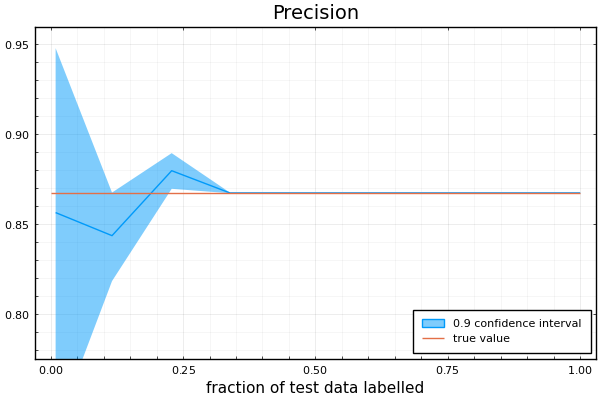}
\includegraphics[width=0.31\textwidth]{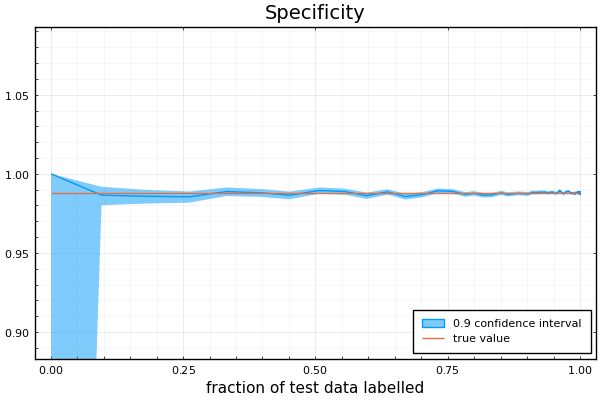}\includegraphics[width=0.3\textwidth]{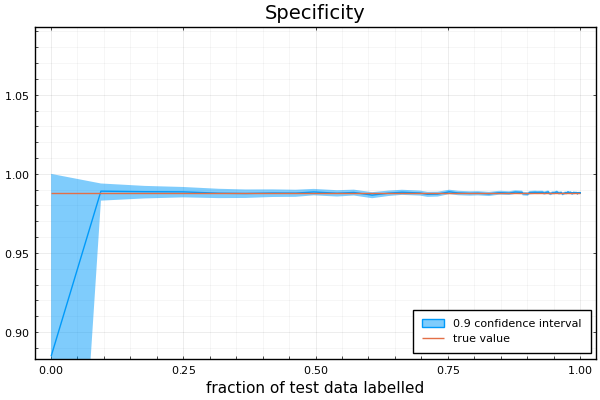}\includegraphics[width=0.31\textwidth]{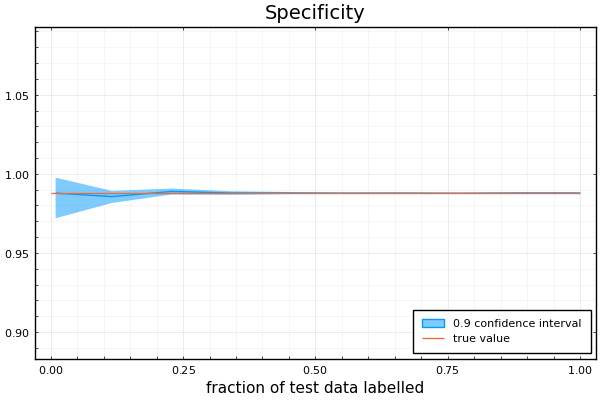}
\caption{Estimating Test Metrics for MNIST. Left: Uniform Sampling. Middle: Importance Sampling. Plotted on the x-axis is the number of distinct samples as a fraction of the test data (not the total number of samples drawn including repetitions). Right: Bernoulli Sampling.\label{fig:mnist:is:bs:app}}
\end{figure}

\begin{figure}
\centering
\includegraphics[width=0.45\textwidth]{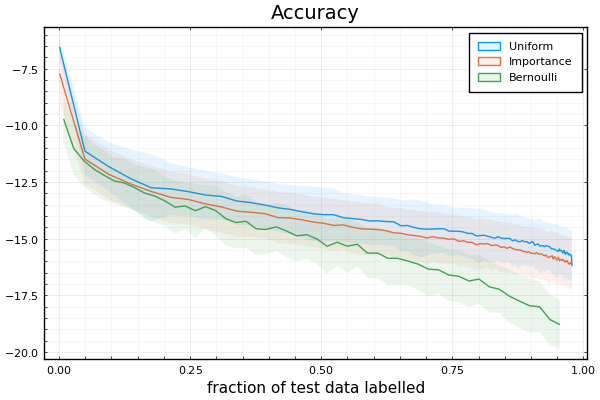}
\includegraphics[width=0.45\textwidth]{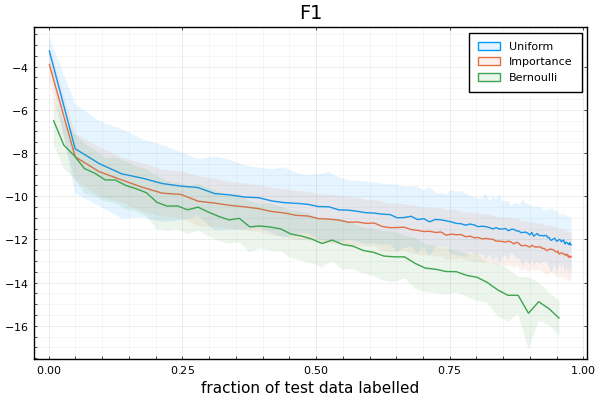}
\includegraphics[width=0.45\textwidth]{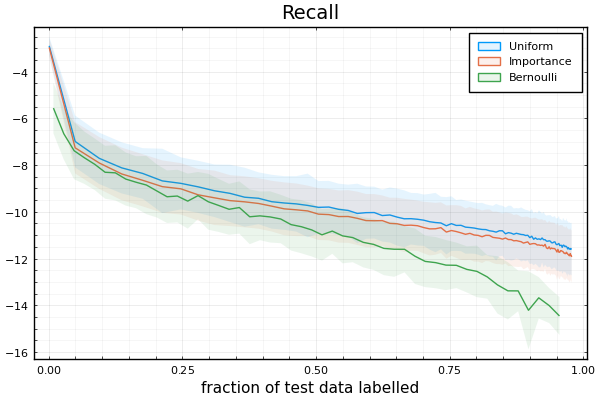}
\includegraphics[width=0.45\textwidth]{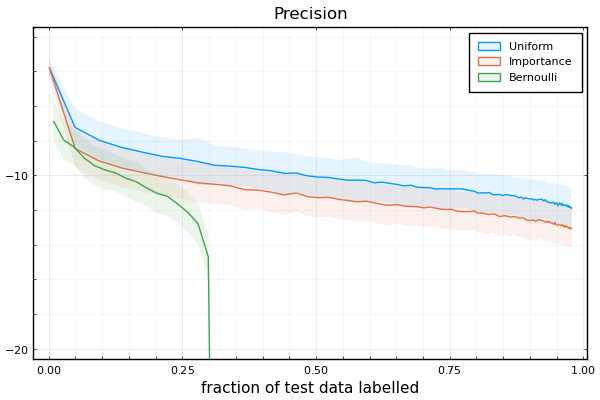}
\includegraphics[width=0.45\textwidth]{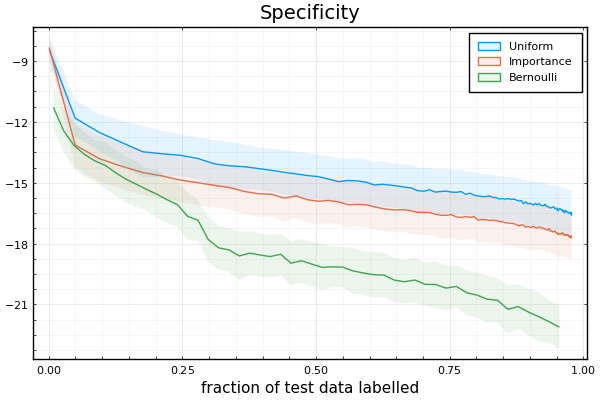}
\caption{Estimating Test Metrics for MNIST. Plotted on the y-axis is the expected log error in estimating each metric: $\log (\text{true}-\text{estimate})^2$ averaged over 3000 experiments, along with 0.5 standard deviation.  Plotted on the x-axis is the number of distinct samples as a fraction of the test data. Optimal Importance Sampling has superior performance compared to simply uniformly selecting test points to label. Optimal Bernoulli Sampling however significantly outperforms both Optimal Importance Sampling and Uniform Random Sampling. \label{fig:mnist:log:us:is:bs:app}}
\end{figure}

\begin{figure}
\centering
\includegraphics[width=0.45\textwidth]{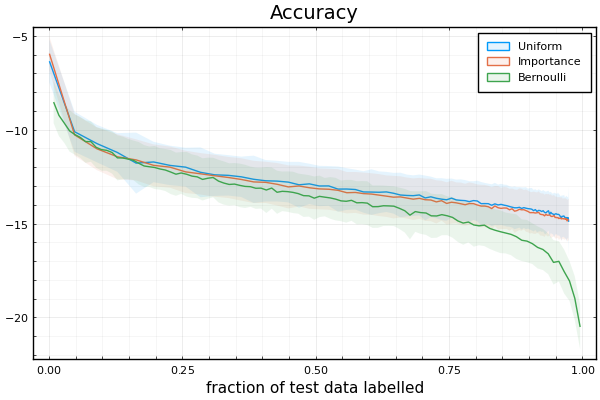}
\includegraphics[width=0.45\textwidth]{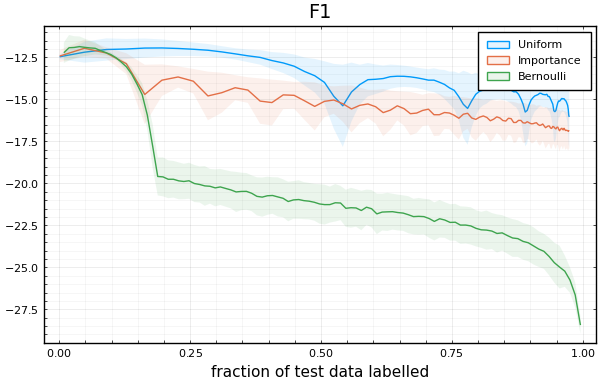}
\includegraphics[width=0.45\textwidth]{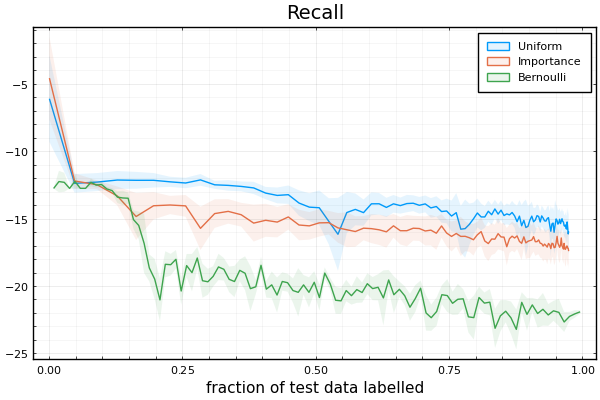}
\includegraphics[width=0.45\textwidth]{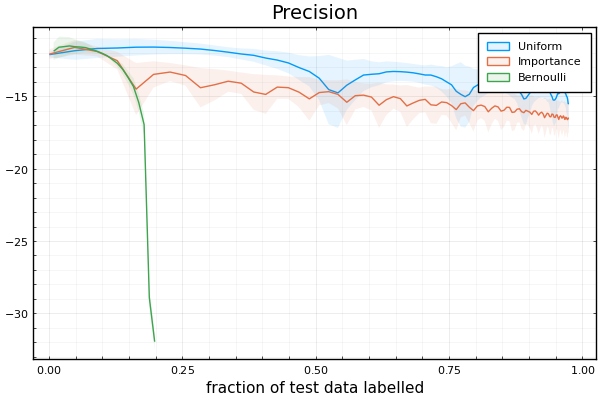}
\includegraphics[width=0.45\textwidth]{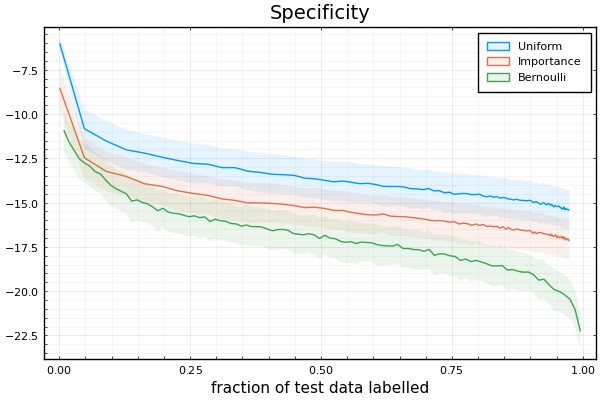}
\caption{Estimating Test Metrics for the Newsgroup classification problem using a Naive Bayes predictor. Plotted on the y-axis is the expected log error in estimating each metric: $\log (\text{true}-\text{estimate})^2$ averaged over 3000 experiments, along with 0.5 standard deviation.  Plotted on the x-axis is the number of distinct samples as a fraction of the test data. Optimal Importance Sampling has superior performance compared to simply uniformly selecting test points to label. Optimal Bernoulli Sampling however significantly outperforms both Optimal Importance Sampling and Uniform Random Sampling. The undulations in the error is a discretisation artefact of the Naive Bayes predictor. \label{fig:mnist:log:us:is:bs:ng:app}}
\end{figure}

\subsection{Cross Results\label{sec:cross:results}}

\begin{figure}
\centering
\includegraphics[width=0.45\textwidth]{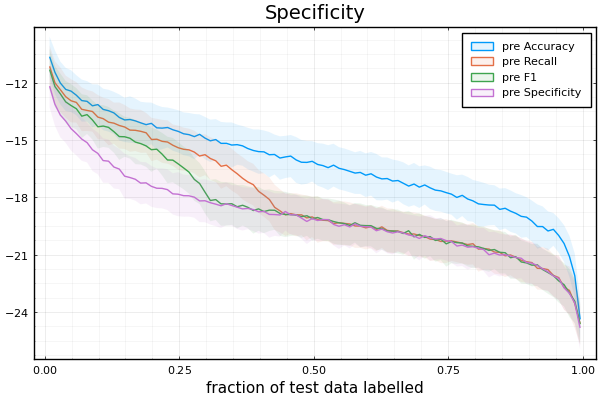}
\includegraphics[width=0.45\textwidth]{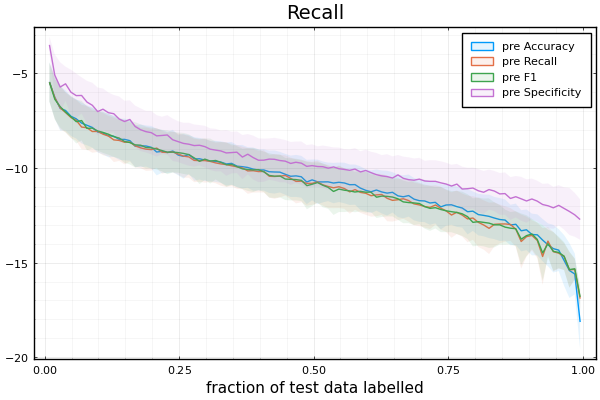}
\includegraphics[width=0.45\textwidth]{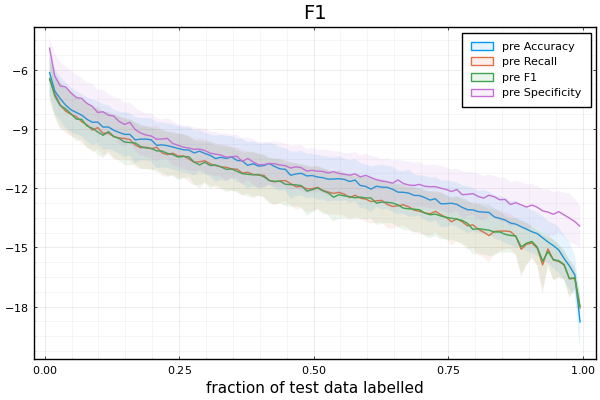}
\includegraphics[width=0.45\textwidth]{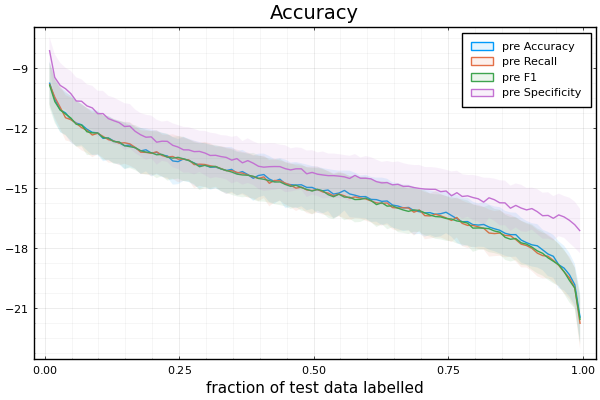}
\caption{Estimating post-sampling Test Metrics using Optimal Bernoulli Sampling for the MNIST classification problem for a variety of pre-sampling metrics (in each legend). Plotted on the y-axis is the expected log error in estimating each metric: $\log (\text{true}-\text{estimate})^2$ averaged over 500 experiments, along with 0.5 standard deviation.  Plotted on the x-axis is the number of distinct samples as a fraction of the test data.  \label{fig:mnist:pre:post:bs}}
\end{figure}

\begin{figure}
\centering
\includegraphics[width=0.45\textwidth]{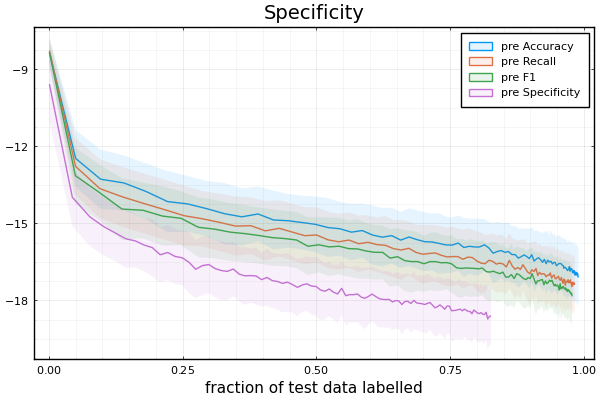}
\includegraphics[width=0.45\textwidth]{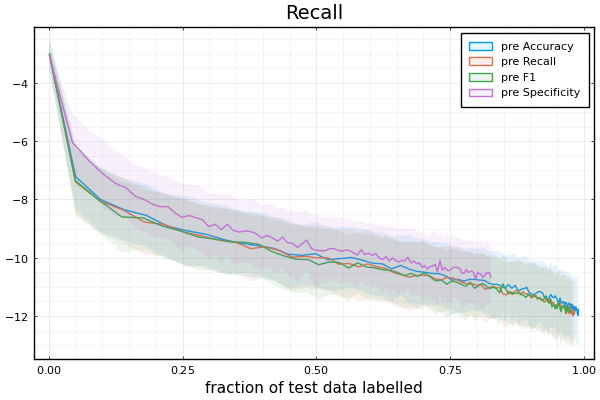}
\includegraphics[width=0.45\textwidth]{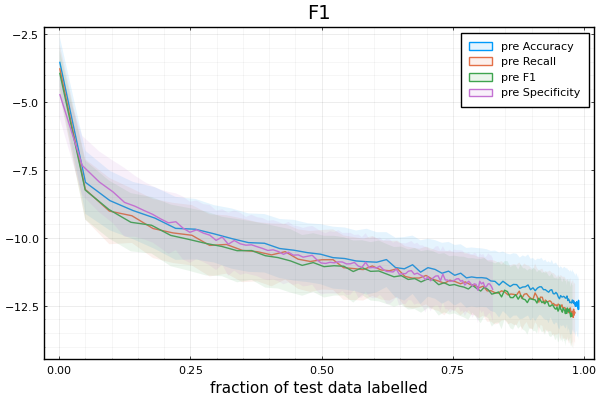}
\includegraphics[width=0.45\textwidth]{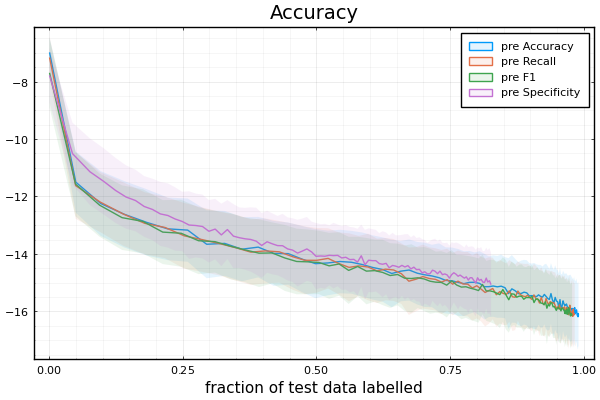}
\caption{Estimating post-sampling Test Metrics using Optimal Importance Sampling for the MNIST classification problem for a variety of pre-sampling metrics (in each legend). Plotted on the y-axis is the expected log error in estimating each metric: $\log (\text{true}-\text{estimate})^2$ averaged over 500 experiments, along with 0.5 standard deviation.  Plotted on the x-axis is the number of distinct samples as a fraction of the test data.  \label{fig:mnist:pre:post:is}}
\end{figure}

We plot here how the choice of pre-sampling metric affects the accuracy of estimating the post-sampling metric. For example, we would expect that if we are interested in the post-sampling metric $F_1$, then the samples that are drawn according to the  pre-sampling metric $F_1$ would be superior than any other pre-sampling metric. We show results for the MNIST problem and the Importance \figref{fig:mnist:pre:post:is} and Bernoulli approaches, \figref{fig:mnist:pre:post:bs}. The results support that if we wish to get the best estimator for a chosen test metric, then it is best to define the optimal sampler using that metric. In other words, for example the errors in estimating Specificity are in general lower when we define the optimal sampler based on Specificity rather than another metric.

\end{document}